\documentclass{article}

\PassOptionsToPackage{numbers, compress}{natbib}

\usepackage[preprint]{neurips_2025}

\usepackage[utf8]{inputenc} 
\usepackage[T1]{fontenc}    
\usepackage{hyperref}       
\usepackage{url}            
\usepackage{booktabs}       
\usepackage{amsfonts}       
\usepackage{nicefrac}       
\usepackage{microtype}      
\usepackage{xcolor}         

\usepackage{amsmath}
\usepackage{enumitem}       
\usepackage[pdftex]{graphicx}
\usepackage{wrapfig}        
\usepackage{subcaption}
\usepackage[capitalise]{cleveref}
\usepackage{bm}
\usepackage{multirow}

\usepackage{tcolorbox}
\usepackage{algorithm}
\usepackage{algpseudocode}

\DeclareMathOperator*{\argmax}{arg\,max}
\DeclareMathOperator*{\argmin}{arg\,min}
\DeclareMathOperator{\diag}{diag}
\DeclareMathOperator{\Tr}{Tr}
\DeclareMathOperator{\cost}{T}
\newcommand{\DTV}{\mathrm{D_{TV}}}
\newcommand{\DKL}{\mathrm{D_{KL}}}
\newcommand{\mat}[1]{\mathbf{#1}}
\newcommand{\vb}[1]{\bm{#1}}
\newcommand{\frobenius}{\mathrm{F}}
\newcommand{\transpose}{\mathsf{T}}

\newcommand{\pyindent}[1]{\hspace*{#1}}
\def\n{\textbackslash{}n}
\def\dtbias{\textit{dt\_bias}}

\usepackage{amsmath,algorithm,tabularx}
\makeatletter
\newcommand{\algmultiline}[1]{%
  \begin{tabularx}{\dimexpr\linewidth-\ALG@thistlm}[t]{@{}X@{}}
    #1
  \end{tabularx}
}
\makeatother

\algnewcommand{\Requirement}[1]{%
  \State \textbf{Require:}
  \Statex \hspace*{\algorithmicindent}\parbox[t]{.8\linewidth}{\raggedright #1}
}

\algnewcommand{\Initialization}[1]{%
  \State \textbf{Initialize:}
  \Statex \hspace*{\algorithmicindent}\parbox[t]{.8\linewidth}{\raggedright #1}
}

\algblock{Scope}{EndScope}
\algtext*{Scope}
\algtext*{EndScope}

\newcounter{figurepair}
\newcommand{\paircaption}[1]{%
  \par 
  \refstepcounter{figurepair}
  \noindent(\arabic{figurepair}) #1
  \par\nobreak 
  \addvspace{\smallskipamount}}%

\title{RAD: Redundancy-Aware Distillation for Hybrid Models via Self-Speculative Decoding}

\author{%
Yuichiro Hoshino \quad 
Hideyuki Tachibana\thanks{Currently with NII LLMC, Tokyo, Japan. Also with Asia University, Tokyo, Japan.} \quad 
Muneyoshi Inahara \quad 
Hiroto Takegawa \\
PKSHA Technology inc.,\\
Bunkyo, Tokyo, Japan \\
}

\begin{document}

\maketitle

\begin{abstract}
    Hybrid models combining Transformers and State Space Models (SSMs) are promising for balancing performance and efficiency. 
    However, optimizing these hybrid models, particularly by addressing the potential redundancy inherent within the Transformer components, remains a significant challenge. 
    In this paper, we propose RAD (Redundancy-Aware Distillation), a novel framework that uses self-speculative decoding as a diagnostic tool to identify redundant attention layers within the model. 
    These identified layers are then selectively replaced with SSM components, followed by targeted (self-)distillation. 
    Specifically, RAD focuses knowledge transfer on the components identified as redundant, considering architectural changes and specific weight initialization strategies. 
    We experimentally demonstrate that self-distillation using RAD significantly surpasses the performance of the original base model on mathematical and coding tasks. 
    Furthermore, RAD is also effective in standard knowledge distillation settings, achieving up to approximately 2x faster convergence compared to baseline methods. 
    Notably, while a baseline model distilled from a Llama-3.1 70B teacher achieves scores of 46.17 on GSM8K and 22.75 on CRUX, RAD achieves significantly higher scores of 71.27 on GSM8K and 28.25 on CRUX, even when using a much smaller Llama-3.1 8B teacher. 
    RAD offers a new pathway for efficient optimization and performance enhancement in the distillation of hybrid models.
\end{abstract}

\section{Introduction}

The Transformer architecture \citep{vaswani2017attention} has become the standard model in many fields, including natural language processing. 
However, its core self-attention mechanism requires computational cost quadratic in the input sequence length, posing a significant bottleneck, especially when dealing with long contexts. 
To address this challenge, recurrent architectures with linear or near-linear complexity, such as State Space Models (SSMs) \citep{gu2021combining, guefficiently, mamba, mamba2}, have garnered attention.
Recently, hybrid models combining the powerful representation capabilities of Transformers with the computational efficiency of SSMs \citep{junxiongdaniele2024mambainllama, bick2024transformers, yang2024gdn, blakeman2025nemotron-h} have emerged, showing promise for achieving new trade-offs between performance and efficiency.
However, when constructing or optimizing efficient hybrid models based on existing powerful pre-trained Transformer models, a key design question arises: which attention layers should be replaced with SSM blocks? 

Previous approaches have often resorted to ad hoc strategies, such as replacing layers at equal intervals \citep{junxiongdaniele2024mambainllama}. While this might be less critical for models trained from scratch, such as Jamba \citep{lieber2024jamba}, Zamba \citep{glorioso2024zamba}, or Samba \citep{ren2024samba}, where the entire architecture is optimized end-to-end, it is plausible that significant functional structures or specialization have emerged across layers in models that have undergone extensive pre-training and instruction tuning. 
Indeed, a growing body of evidence indicates that not all parameters or layers within Transformer models contribute equally to their overall performance \citep{dalvi-etal-2020-analyzing, NEURIPS2019_2c601ad9, bian-etal-2021-attention}. Some studies have shown that removing certain attention layers can result in surprisingly minor performance degradation on specific tasks \citep{men2024shortgpt, gromov2024unreasonable, he2024matterstransformersattentionneeded}, suggesting that a degree of layer redundancy is inherent. 
Furthermore, distinct functional roles across different Transformer layers have also been observed \citep{dong2024painters}.

Motivated by this background, we hypothesize that some computationally expensive attention layers might be redundant relative to their contribution to the overall model performance. 
We posit that by \textbf{identifying} these potentially redundant layers and \textbf{selectively} replacing them with SSMs, we can achieve more efficient and effective model optimization, particularly through knowledge distillation \citep{hinton2015distilling}. 
Unlike ad hoc layer selection, considering the model-specific redundancy aims to enhance computational efficiency while maintaining or even improving performance.

In this paper, we propose \textbf{RAD (Redundancy-Aware Distillation)} to tackle this challenge. 
RAD is a novel framework that repurposes the mechanism of \textbf{Self-Speculative Decoding} \citep{leviathan2023fast, chen2023accelerating} from its original goal of accelerating inference to serve as a \textbf{diagnostic tool for identifying the redundancy of attention layers} within a model. 
First, RAD identifies attention layers with high redundancy within a given Transformer model. 
Second, it selectively replaces these identified layers with efficient SSM blocks to construct a hybrid model. 
Finally, it distills knowledge from the original model (or a larger teacher model) to the constructed hybrid student model. 
This knowledge transfer is ``Redundancy-Aware'', focusing on the replaced components and considering architectural changes and effective weight initialization strategies.
Our main contributions are as follows:
\begin{itemize}[leftmargin=2em]
    \item \textbf{A Novel Application of Self-Speculative Decoding:} We propose using it as a diagnostic tool to identify layer-level computational redundancy in models.
    \item \textbf{The RAD Framework:} We introduce a systematic methodology for constructing hybrid models by considering redundancy and performing targeted, efficient (self-)distillation.
    \item \textbf{Performance Improvement in Hybrid Models:} We experimentally demonstrate that self-distillation using RAD can lead to a ``Born-Again''-like effect \citep{pmlr-v80-furlanello18a}, where the student model outperforms the original teacher on reasoning tasks like math and coding.
    \item \textbf{Effectiveness in Standard Distillation:} We show that the RAD framework is also effective in standard knowledge distillation setups, achieving significantly faster convergence compared to baseline methods and yielding accuracy that surpasses even distillation from a much larger teacher model.
\end{itemize}

\section{Preliminaries}
\label{sec:background}

In this section, we explain the main technical background necessary to understand our proposed method, RAD (Redundancy-Aware Distillation). 
We outline the basic concepts of speculative decoding, State Space Models (SSMs), and Hybrid models.

\subsection{Speculative Decoding}
\label{subsec:speculative_decoding}

Speculative decoding \citep{leviathan2023fast, chen2023accelerating} is a technique to accelerate inference from large autoregressive models (target model, denoted as $\mathcal{M}_p$). 
In vanilla autoregressive models,
decoding $K$ tokens typically requires $K$ sequential model runs, which is time-consuming. 
The basic idea of speculative decoding is to first use a computationally cheaper (i.e., faster) draft model $M_q$ to speculatively generate multiple candidate tokens. 
Subsequently, these candidate tokens and their probabilities are verified \textit{in parallel} using the original target model $\mathcal{M}_p$.
The fundamental goal is thus to minimize the number of computationally expensive forward passes required by the target model $\mathcal{M}_p$ for generating a sequence, thereby accelerating the overall inference process.
The core of this verification process is a sampling method called speculative sampling, which crucially ensures that the output distribution remains identical to that of the target model $\mathcal{M}_p$ alone. 
The algorithm is described in Supplementary Materials.

\subsection{State Space Models (SSMs)}
\label{subsec:ssm}

SSMs \citep{gu2021combining,guefficiently} have gained attention as a promising alternative architecture to address the computational complexity issues of the self-attention mechanism in Transformers. 
Originating from classical control theory, SSMs compress sequence history into a fixed-size hidden state $\vb{h}_t$, assuming this state evolves linearly over time. 
In discrete form, this is often represented by the recurrence relations:
\begin{align*}
\vb{h}_t &= \mat{\bar{A}}\vb{h}_{t-1} + \mat{\bar{B}}\vb{x}_t \\
\vb{y}_t &= \mat{C} \vb{h}_t + \mat{D} \vb{x}_t
\end{align*}
where $\vb{x}_t$ is the input, $\vb{y}_t$ is the output, $\vb{h}_t$ is the hidden state, and $\mat{\bar{A}}, \mat{\bar{B}}, \mat{C}, \mat{D}$ are system matrices. 
Key advantages of SSMs include linear $O(L)$ computational complexity with respect to sequence length $L$, and high memory efficiency during inference, as only a constant-size hidden state needs to be maintained, in contrast to the $O(Ld)$ KV cache growth in Transformers.
On the other hand, it has been discussed that SSMs also have challenges of potential limitation of memory capacity \citep{yang2024deltanet, 10.5555/3692070.3692933, chen2024stuffedmambastatecollapse} due to the fixed size of state vector.

Among SSMs, Mamba \citep{mamba} is particularly prominent recently.
Built upon Structured State Space models (S4), Mamba incorporates the selection mechanism that dynamically controls SSM parameters ($\mat{\bar{A}}, \mat{\bar{B}}$, and the step size $\Delta$) depending on the current input $\vb{x}_t$.
Mamba also employs hardware-aware parallel algorithms (selective scan) optimized for GPU memory hierarchy. 
Mamba2 \citep{mamba2} further refines Mamba's selective SSM layer based on the theory of State Space Duality (SSD), achieving speedups over Mamba-1 while maintaining high performance. 

Importantly, the recurrent form of SSMs could be understood as implicit solvers of online learning problems \citep{longhorn2024}, or test-time regressors \citep{wang2025testtimeregressionunifyingframework}.
In other words, SSMs perform online associative recall at test time. 
This implicit link provides a theoretical basis for SSM design, naturally incorporating mechanisms for forgetting and updating. 
In particular, a mechanism called Longhorn \citep{longhorn2024} has been reported to exhibit excellent sample efficiency and extrapolation capabilities to long contexts compared to Mamba.

\subsection{Hybrid Models}
\label{subsec:hybrid_models}
Hybrid architectures, which combine components from Transformers and SSMs, are actively researched to leverage the distinct strengths of each: the precise attention capabilities of Transformers and the efficient long-context processing of SSMs, thereby aiming for a superior balance between performance and efficiency.
Various strategies for constructing such hybrids have been explored. These include interleaving Transformer and SSM blocks \citep{lieber2024jamba, glorioso2024zamba}, combining windowed attention with Mamba \citep{ren2024samba}, integrating DeltaNet layers with attention components \citep{yang2024deltanet, yang2024gdn}, and replacing attention layers in established Transformer models with SSM blocks, often followed by knowledge distillation to transfer attentional patterns or align intermediate states \citep{junxiongdaniele2024mambainllama, bick2024transformers, goldstein2025radladsrapidattentiondistillation}.
The prevalence and success of these hybrid approaches suggest that while pure SSMs offer significant efficiency gains, they may not yet fully replicate all desirable properties of Transformers alone, making hybridization a practical pathway towards achieving both high performance and enhanced efficiency.

\section{Proposed Method: RAD}
\label{sec:proposed_method}

\begin{figure}
    \centering
    \includegraphics[width=0.8\linewidth]{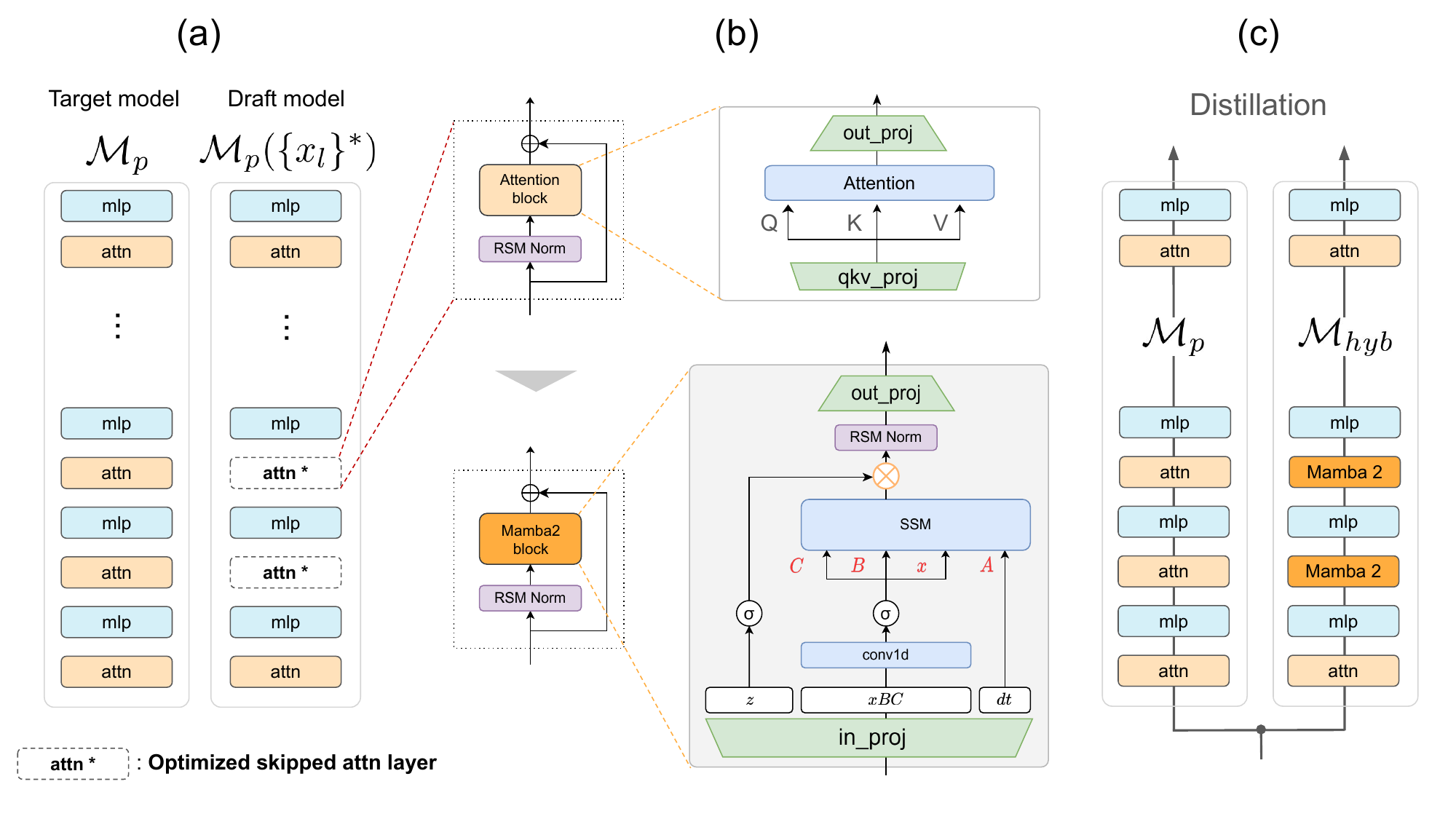} 
    \caption{Overview of our proposed RAD (Redundancy-Aware Distillation) framework. 
    (a) \textbf{Redundancy Identification:} Redundant attention layers are identified via self-speculative decoding (selectively skipping attention layers) and Bayesian Optimization aimed at maximizing the resulting inference throughput. 
    (b) \textbf{Hybrid Model Initialization:} The identified redundant attention layers are replaced with SSM blocks (e.g., Mamba2) using specific weight initialization strategies (copying `out\_proj` weights from the attention block and zero-initializing of `in\_proj` weights) to create the initial hybrid model $\mathcal{M}_{\textit{hyb}}$. 
    (c) \textbf{Redundancy-Aware Distillation:} Knowledge is distilled from the teacher model $\mathcal{M}_{p}$ to the student hybrid model $\mathcal{M}_{\textit{hyb}}$ by training \textit{only} the parameters of the newly added SSM blocks, using forward KL divergence on the final output logits.}
    \label{fig:rad}
\end{figure}

In this work, we propose RAD (Redundancy-Aware Distillation, \cref{fig:rad}), a novel framework to identify computational redundancy within a pre-trained Transformer model ($\mathcal{M}_p$) and leverage this information to construct and optimize an efficient and high-performing hybrid model. 
RAD primarily consists of the following three steps: 
(1) Identifying redundant attention layers using self-speculative decoding and Bayesian Optimization, 
(2) Constructing the hybrid model by replacing the identified layers with SSM blocks, and 
(3) Performing redundancy-aware (self-)distillation on the constructed hybrid model.

\subsection{Identifying Redundant Attention Layers}
\label{subsec:identify_redundancy}

The core idea of RAD is to use the inference speedup gained from \textbf{Self-Speculative Decoding} as a proxy for identifying computationally redundant attention layers. 
Specifically, following the approach of \citep{zhang2023draft}, we construct a faster draft model by skipping specific attention layers of the target model $\mathcal{M}_p$ during computation. Let us denote it as $\mathcal{M}_p(\{x_l\}_{l=1}^L)$, where $\{x_l\}_{l=1}^L$ is an array of binary variables to specify which attention layers to skip, where $x_l=1$ means skipping the $l$-th attention layer of $\mathcal{M}_p$ when constructing the draft model, and $x_l=0$ means keeping it. 

We then measure the \textbf{inference throughput} ($\tau$ tokens/sec) achieved when performing self-speculative decoding using this draft model and the original target model $\mathcal{M}_p$. 
Note that the throughput $\tau$ is not merely a function of the number of skipped attention layers. Instead, it heavily depends on the similarity between the teacher $\mathcal{M}_p$ and the draft $\mathcal{M}_p(\{x_l\})$.
As the draft model is required to propose more candidate samples if its proposal is rejected by the teacher model $\mathcal{M}_p$, the throughput $\tau$ gets smaller as the original $\mathcal{M}_p$ and the draft $\mathcal{M}_p(\{x_l\})$ get dissimilar.
In other words, higher throughput $\tau$ is achieved when the draft model well approximates the original model, as well as when many attention layers are pruned.
\emph{We hypothesize that the attention layers whose removal (skipping) leads to the greatest increase in generation speed (tokens/sec), while maintaining reasonable accuracy (implicitly captured by the token acceptance rate), are the most computationally redundant.}

Based on the observations above, we can hypothesize that optimizing throughput $\tau$ leads a good identification of redundant attention layers.
Let $\tau(\mathcal{M}_p, \{x_l\})$
denote the inference throughput using a draft model constructed by removing $\{x_l\}$ layers from the teacher Transformer $\mathcal{M}_p$.
Now, our current goal is to find the best skip configuration $\{x_l\}^*$ that maximizes the expected inference throughput $\mathbb{E}[\tau(\mathcal{M}_p, \{x_l\})]$:
\begin{align}
\{x_l\}^{*} = \argmax_{\{x_l\} \in \{0,1\}^L} \mathbb{E}[\tau (\mathcal{M}_p, \{x_l\})]. \label{eq:discrete_opt_rad_en}
\end{align}
Since exhaustively exploring all $2^L$ possible skip configurations is computationally infeasible for typical values of $L$ (e.g., $L=28$ for Llama-3.2-3B), and the relationship between $\{x_l\}$ and $\tau$ is complex and likely non-convex, we employ Bayesian Optimization (BO) to efficiently search for an effective skip configuration
\footnote{
Details of our BO procedure, including the adaptation for discrete search spaces based on \citep{zhang2023draft}, are provided in Supplementary Materials.
}.
After running the BO procedure for a sufficiently large number of iterations, we select the binary configuration $\{x_l\}^* \in \{0,1\}^L$ that yielded the best observed objective value during the search. 
The attention layers for which $x_l^*=1$ are identified as computationally redundant (i.e., likely to maximize inference throughput if skipped).

\subsection{Hybrid Model Construction / Initialization}
\label{subsec:hybrid_construction}

Once the redundant attention layers are identified, the next step is to construct the hybrid model $\mathcal{M}_{\textit{hyb}}$ based on the original Transformer model $\mathcal{M}_p$.
Specifically, based on the skip configuration $\{x_l\}^*$ determined in Section \ref{subsec:identify_redundancy}, we replace the attention layers where $x_l^*=1$ with chosen SSM blocks. 
During replacement, architectural adjustments are made, such as ensuring consistency in the input/output dimensions between the original attention block and the new SSM block.
The initialization of the newly added SSM blocks is crucial for stable and efficient learning. We adopt the following strategy:
\begin{itemize}[leftmargin=*]
    \item \textbf{Initialization of Out Projection Layer:} Following prior work \citep{junxiongdaniele2024mambainllama}, we utilize the weights of the `out\_proj' layer from the original attention block being replaced as the initial values for the `out\_proj' layer of the newly introduced SSM block.
    \item \textbf{Initialization of Internal SSM Weights (Replicating Skipped State):} This point is particularly crucial. Without this initialization, the model's performance degrades dramatically, and the loss also does not decrease readily (details are described in Supplementary Materials). We aim to replicate the state where the attention layer was skipped during self-speculative decoding as the initial state at the beginning of distillation. 
    Specifically, to ensure the output from the SSM block is initially zero (mimicking the skipped state), we initialize a part of the `in\_proj' layer weights of the SSM block with zeros. 
    This ensures that the replaced SSM block, immediately after replacement, behaves similarly to a skipped attention layer by not passing information through initially.
\end{itemize}

\subsection{Redundancy-Aware (Self-)Distillation}
\label{subsec:rad_distillation}

The final step is to distill knowledge from the original model $\mathcal{M}_p$ (teacher) to the constructed hybrid model $\mathcal{M}_{\textit{hyb}}$ (student) to further enhance its performance. While we term this process Redundancy-Aware, the specific strategy we adopt is as follows:

\paragraph{Selective Parameter Training.} During the distillation process, instead of updating the entire model's parameters, we \textit{only train the parameters of the SSM blocks that were replaced} in the step described in \cref{subsec:hybrid_construction}.
The parameters of the original Transformer blocks remain frozen. This focuses the learning effort on optimizing the newly introduced efficient components identified via redundancy analysis.

\paragraph{Loss Function.} For knowledge distillation, we \textbf{\textit{solely}} use the forward Kullback-Leibler (KL) divergence between the final output logits of the teacher model ($\mathcal{M}_p$) and the student model ($\mathcal{M}_{\textit{hyb}}$):
\begin{equation}
\mathcal{L}_{\text{KD}} = D_{\text{KL}}( P_{\text{teacher}} \;\|\; P_{\text{student}} )
\label{eq:kl_loss_en}
\end{equation}
Here, $P_{\text{teacher}}$ and $P_{\text{student}}$ represent the output probability distributions from the teacher and student models, respectively.

This distillation process is primarily performed in the context of Self-Distillation (learning from the teacher $\mathcal{M}_p$ from which the student $\mathcal{M}_{\textit{hyb}}$ was derived).
Additionally, this RAD framework is also applicable to standard knowledge distillation scenarios using a larger external model as the teacher. 
In this study, we also report its effectiveness in that setting (where all parameters of the student model $\mathcal{M}_{\textit{hyb}}$ are trainable) in Section \ref{subsec:results}.

\section{Experiments and Results}
\label{sec:experiments}

In this section, we describe the experimental setup designed to evaluate the effectiveness of our proposed RAD framework and present the main results.
Throughout our experiments, we use the Llama3.2-3B-Instruct as the base pre-trained Transformer model ($\mathcal{M}_p$) from which we identify redundancy and construct hybrid models.

\subsection{Experimental Setup}
\label{subsec:setup}

\paragraph{Redundancy Identification Setup.} To identify redundant attention layers (Section \ref{subsec:identify_redundancy}), we randomly select 10 instances from the PG-19 dataset \citep{raecompressive2019}. 
For each instance, we perform zero-shot summarization which requires global information of the whole context \citep{sun2024triforce}. 
We use the initial part of the book text as input, truncated to 2048 tokens, and generate 128 tokens using the self-speculative decoding (greedy setting) with different layer skipping configurations $\{x_l\}$. 
The average throughput (tokens/sec) over these 10 instances is used as the objective function $\mathbb{E}[\tau(\mathcal{M}_p, \{x_l\})]$ for Bayesian Optimization \citep{bayesianoptimization}.
We run the Bayesian Optimization for 2000 iterations, which takes several hours on an H100 GPU.

\paragraph{Hybrid Model Configuration and Training.} Based on the results from the redundancy identification step, we construct the hybrid models $\mathcal{M}_{\textit{hyb}}$ by replacing the identified attention layers with SSM blocks (Mamba \citep{mamba}, Mamba2 \citep{mamba2}, or Longhorn \citep{longhorn2024}). 
Weight initialization follows the strategy described in Section \ref{subsec:hybrid_construction} (copying `out\_proj`, zero-initializing of `in\_proj`).
The training settings for the hybrid models are as follows (training takes 2-3 days on 8×A100 GPUs):

\begin{description}[leftmargin=2ex]
\item[\emph{\underline{Self-Distillation \textup(partial distillation with self-supervision\textup)}}] Using the base model Llama3.2-3B-Instruct as the teacher $\mathcal{M}_p$, we train \textbf{only the parameters of the replaced SSM blocks,} keeping other parameters frozen, as described in Section \ref{subsec:rad_distillation}. We use the AdamW optimizer with a cosine learning rate schedule. For Mamba and Longhorn, this schedule includes a peak learning rate of 2e-4 decaying to a minimum of 2e-5. 
For Mamba2, the peak is 2e-5, and the decay proceeds without a specified minimum learning rate.

\item[\emph{\underline{Standard Distillation \textup(full-parameter distillation from large teacher models\textup)}}] 
Using a larger external model, Llama-3.1-8B-Instruct, as the teacher, we train \textbf{all parameters} of the student model $\mathcal{M}_{\textit{hyb}}$. 
We use the AdamW optimizer with a learning rate of 2e-5, following a cosine decay schedule.
For comparison, we use publicly available distilled models provided by \citep{junxiongdaniele2024mambainllama}
as baselines.

Specifically, we use the \texttt{Llama3.2-Mamba-3B-distill} (50\% attn, equal intervals)\footnote{\url{https://huggingface.co/JunxiongWang/Llama3.2-Mamba-3B-distill}} and 
\texttt{Llama3.2-Mamba2-3B-distill} (50\% attn, equal intervals)\footnote{\url{https://huggingface.co/JunxiongWang/Llama3.2-Mamba2-3B-distill}} which were distilled from the Llama-3.1-70B-Instruct.
These models were created by replacing 50\% of the attention layers in Llama3.2-3B-Instruct with Mamba or Mamba2 blocks at equal intervals, 
initializing them with the weights from the original attention block ('qkv\_proj' and 'out\_proj' in \cref{fig:rad}), and then performing standard knowledge distillation using KL divergence loss only.

For a fair comparison in the standard distillation setup, we constrained our redundancy identification optimization \cref{eq:discrete_opt_rad_en} to select the same number of layers for replacement as the baselines (i.e., 50\% of total attention layers).
Details are described in Supplementary Materials.
\end{description}

For both distillation setups, we set context length during distillation to 2048 tokens and 
train for one epoch on a dataset mixture derived from GenQA \citep{genqa2024}, InfinityInstruct \citep{InfinityInstruct2024}, and OpenHermes 2.5 \citep{OpenHermes2.5}, totaling approximately 20B tokens, to ensure comparability with the baseline setup.

\paragraph{Evaluation Metrics.} We evaluate model performance across a wide range of tasks to assess various capabilities, including reasoning and long-context processing. 
For general benchmarks, we evaluate 10 tasks using the LM Evaluation Harness library \citep{eval-harness}: WinoGrande (WG), PIQA (PI), HellaSwag (HS), ARC-Easy (AE), ARC-Challenge (AC), MMLU (MM), OpenBookQA (OB), TruthfulQA (TQ), PubMedQA (PM), and RACE (RA). Additionally, we assess reasoning capabilities in a zero-shot setting using GSM8K \citep{cobbe2021gsm8k} and CRUX \citep{gu2024cruxeval}. 
Long-context retrieval ability is tested using the Passkey retrieval, 
which involves finding a 7-digit number embedded within varying positions in Paul Graham's essays. 
Furthermore, we evaluate performance on diverse long-context tasks using the LongBench \citep{bai2024longbench}, comprising 15 datasets:
Single.\ QA (NarrativeQA, Qasper, MultiFieldQA), 
Multi.\ QA (HotpotQA, 2WikiMQA, Musique), 
Summ.\ (GovReport, QMSum, MultiNews), 
Few-shot (TREC, TriviaQA, SAMSum), 
Psg ret.\ (PassageRetrieval) and Code (LCC, RepoBench-P). 
The evaluation code utilized is primarily from \citep{junxiongdaniele2024mambainllama} and \citep{yuan_liu_zhong_2024_kvcache_comp_benchmnark}.

\subsection{Main Results}
\label{subsec:results}

\paragraph{\emph{\underline{Self-Distillation.}}}
Our redundancy identification step (\cref{subsec:identify_redundancy}) yielded a configuration where 8 specific attention layers were identified as optimal for replacement to maximize inference throughput.
All models reported in this subsection were evaluated using greedy decoding. 
\textbf{In the following tables, bold indicates the best performing model for each task, and underline indicates the second best.}

\paragraph{Performance on General Language Modeling.}
Table \ref{tab:lm_eval_self_distill} presents the performance of the RAD models on 10 tasks from the LM Evaluation Harness and their average scores. The base Llama-3.2-3B-Instruct model is included for comparison. The RAD models demonstrated competitive performance across these general benchmarks against the teacher model. Notably, on several individual tasks, the RAD models were also observed to outperform the teacher model.

\begin{table}[t]
    \centering
    \caption{Evaluation on LM Eval benchmark for RAD self-distilled models. The base model Llama-3.2-3B-Instruct is the teacher. ``w/ Mamba'', ``w/ Mamba2'', and ``w/ Longhorn'' denote RAD models where 8 identified redundant attention layers are replaced by the respective SSM blocks and then self-distilled.
    (acc, acc\_n denote accuracy and normalized accuracy, respectively.)
    }
    \label{tab:lm_eval_self_distill}
    \setlength{\tabcolsep}{3pt}
    \begin{tabular}{@{}l *{11}{c}@{}}
        \toprule
        Model & WG & PI & HS & AE & AC & MM & OB & TQ & PM & RA & \textbf{Avg.} \\
        & acc & acc & acc\_n & acc & acc\_n & acc & acc\_n & acc & acc & acc & \\
        \midrule
        \multicolumn{12}{@{}l}{\textit{Teacher model}} \\
        Llama-3.2-3B-Ins  & \textbf{67.40} & \textbf{75.79} & \textbf{70.45} & \textbf{74.12} & \textbf{46.08} & \textbf{60.30} & 35.80 & 43.96 & \underline{69.80} & 40.77 & \textbf{58.45} \\
        \midrule
        \multicolumn{12}{@{}l}{\textit{Self-Distilled (8 layers replaced)}} \\
        w/ Mamba          & \underline{66.93} & \underline{74.86} & \underline{69.17} & 73.15 & 44.28 & \underline{58.28} & \underline{37.60} & \textbf{44.47} & \textbf{71.00} & 40.86 & \underline{58.06} \\
        w/ Mamba2         & 65.75 & 74.59 & 68.23 & 72.77 & 43.69 & 55.17 & \textbf{38.20} & 44.12 & 69.60 & \textbf{41.63} & 57.38 \\
        w/ Longhorn       & 66.61 & 74.70 & 69.16 & \underline{73.32} & \underline{45.65} & 58.27 & 37.20 & \underline{44.45} & 69.40 & \underline{41.34} & 58.01 \\
        \bottomrule
    \end{tabular}
\end{table}

\begin{wraptable}[10]{r}{0.53\textwidth}
    \centering
    \vspace{-1.5\baselineskip}
    \caption{Performance on reasoning tasks (0-shot).}
    \label{tab:reasoning_self_distill}
    \setlength{\tabcolsep}{4pt} 
    \begin{tabular}{@{}lcc@{}}
        \toprule
        Model                 & GSM8K                 & CRUX  \\
        \midrule
        Llama-3.2-3B-Ins      & 56.25 (1.0)        & 26.12 (1.0)  \\
        \midrule
        \multicolumn{3}{@{}l}{\textit{Self-Distilled (8 layers replaced)}} \\
        w/ Mamba              & 62.77 (0.95)        & \underline{27.38} (1.44) \\
        w/ Mamba2             & \textbf{64.22} (0.93) & 26.62 (1.55)  \\
        w/ Longhorn           & \underline{63.31} (0.95)    & \textbf{27.62} (1.53)   \\
        \bottomrule
    \end{tabular}
\end{wraptable}
\paragraph{Performance on Reasoning Tasks.}
The performance on mathematical (GSM8K) and code (CRUX) reasoning tasks is detailed in Table \ref{tab:reasoning_self_distill}.
The values in parentheses indicate the average reasoning length ratio relative to the teacher model (teacher = 1.0).
All models are evaluated with ZeroEval \citep{Lin-ZeroEval-A-Unified-2024}.
Here, the RAD models demonstrate a significant improvement over the base Llama-3.2-3B-Instruct model.
These results suggest that our RAD framework, by identifying and replacing specific attention layers deemed redundant and then applying targeted self-distillation, can notably enhance the reasoning capabilities of the base model. 
The significant gains on GSM8K and CRUX, particularly the effect where the student models (RAD variants) outperform their teacher akin to ``Born-Again'' effects \citep{pmlr-v80-furlanello18a}, highlight the effectiveness of RAD in reallocating model capacity by replacing less critical attention layers with efficient SSMs and fine-tuning these new components.
This targeted architectural modification and distillation appear particularly beneficial for tasks requiring complex reasoning.
This is also confirmed in the code completion task in LongBench (the full results are provided in the Supplementary Materials).

\begin{figure}[htbp]
    \centering
    \begin{minipage}[b]{0.48\linewidth}
        \centering
        \includegraphics[scale=0.32]{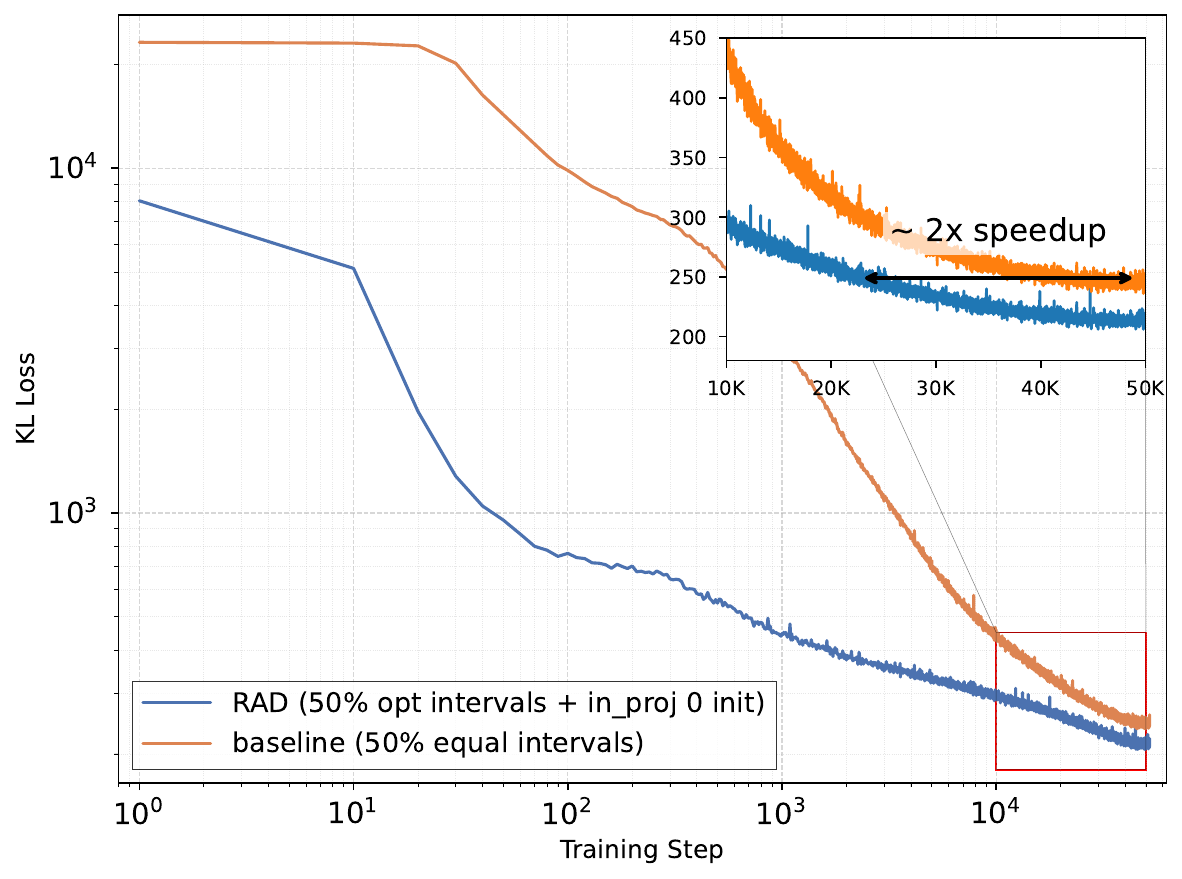} 
        \caption{Comparison of training loss curves with baseline proposed in \citep{junxiongdaniele2024mambainllama} for 50\% Mamba2 distillation.}
        \label{fig:mamba2_distill_loss}
    \end{minipage}
    \hfill 
    \begin{minipage}[b]{0.48\linewidth}
        \centering
        \includegraphics[scale=0.32]{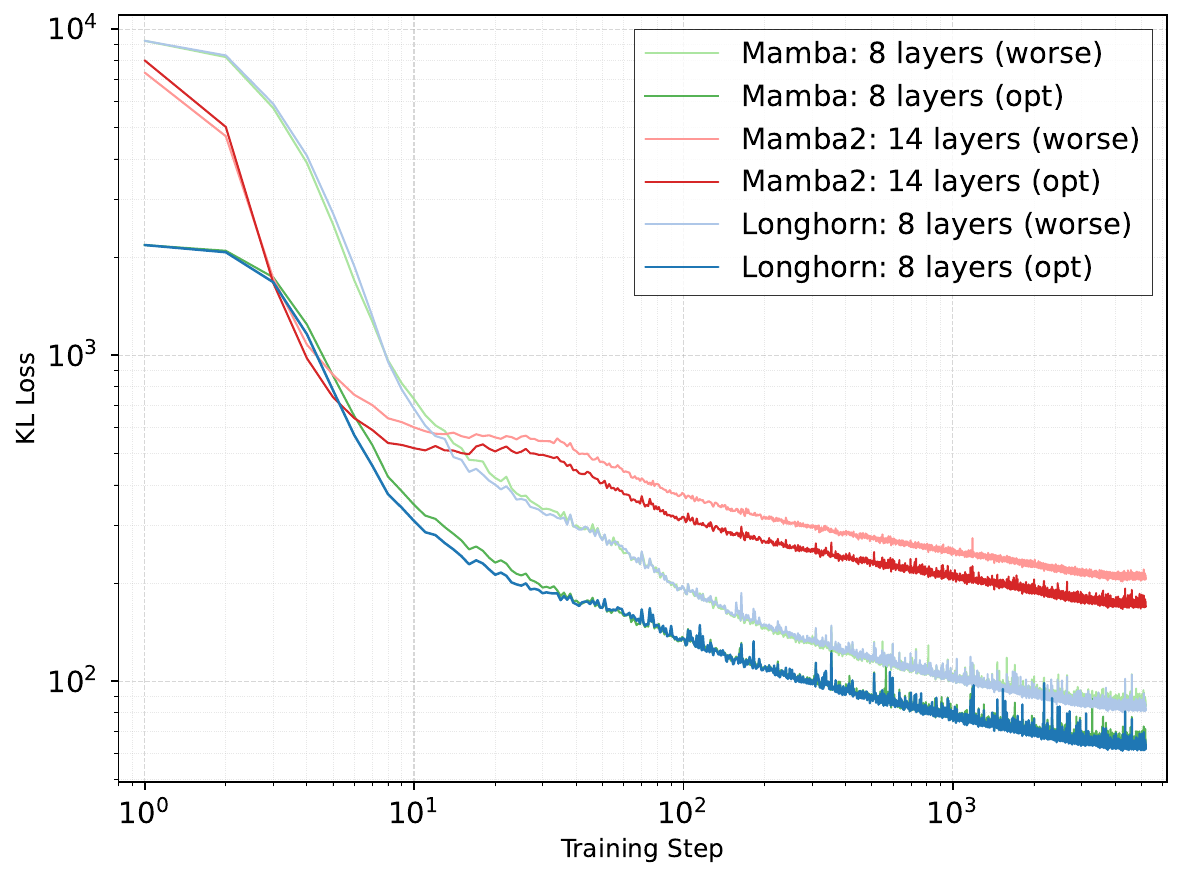} 
        \caption{Comparison of training loss curves for self-distillation setting.
        \textbf{opt}: high throughput attn. layers, \textbf{worse}: low throughput attn. layers.
        }
        \label{fig:self_distill_loss}
    \end{minipage}
\end{figure}

\paragraph{\emph{\underline{Standard Distillation.}}}
In the self-distillation setup, the initialization strategy aimed to replicate the skipped state from self-speculative decoding, which proved beneficial.
In contrast, for standard distillation, where the teacher model is distinct from the student's origin, the direct benefits of such an "optimized" initial state are not inherently guaranteed.
Nevertheless, our approach demonstrates a notable advantage: the training loss decreases rapidly in the early stages. 
This leads to convergence speeds that can be more than twice as fast as baseline methods (as illustrated in \cref{fig:mamba2_distill_loss} for Mamba2).

\cref{tab:lm_eval_distill} presents the LM Evaluation Harness results for the standard distillation setting. 
Our RAD models (denoted as "opt / 8B"), despite being distilled from a smaller Llama-3.1-8B-Instruct teacher, consistently outperform or perform comparably to their respective baseline counterparts ("eql / 70B") that were distilled from a much larger Llama-3.1-70B-Instruct teacher. 

Performance on long-context understanding tasks, evaluated using LongBench and the Passkey Retrieval, is summarized in \cref{tab:longbench_distill}. 
RAD Mamba2 (opt / 8B) demonstrates strong improvements, outperforming its 70B-distilled baseline (Mamba2 (eql / 70B)) on the LongBench average (38.98 vs. 35.0) and the Needle task (74.3 vs. 56.3). While RAD Mamba (opt / 8B) is competitive on the LongBench average, the Mamba (eql / 70B) baseline shows stronger performance on the Needle task in this comparison.

\begin{table}[htb]
    \centering
    \caption{Evaluation on LM Eval benchmark for standard distillation setting. 
    50\% of the attention layers are replaced with Mamba/Mamba2 blocks at equal intervals (eql) or optimized intervals (opt) based on the redundancy identification step, i.e, our RAD model.
    Each model is distilled from the Llama-3.1-70B-Instruct (70B) or Llama-3.1-8B-Instruct (8B) teacher model.
    }
    \label{tab:lm_eval_distill}
    \setlength{\tabcolsep}{3pt}
    \begin{tabular}{@{}l *{11}{c}@{}}
        \toprule
        Model & WG & PI & HS & AE & AC & MM & OB & TQ & PM & RA & \textbf{Avg.} \\
        & acc & acc & acc\_n & acc & acc\_n & acc & acc\_n & acc & acc & acc & \\
        \midrule
        Mamba (eql / 70B)       & 67.72 & \textbf{77.31} & \textbf{70.29} & 77.69 & \textbf{48.38} & \textbf{54.34} & 39.40 & 42.06 & 65.60 & 40.19 & 58.30 \\
        Mamba (opt / 8B)        & \textbf{68.27} & 76.28 & 70.11 & \textbf{78.32} & 47.35 & 53.91 & \textbf{40.0} & \textbf{43.46} & \textbf{69.60} & \textbf{40.77} & \textbf{58.81} \\
        \midrule
        Mamba2 (eql / 70B)      & 66.14 & 75.95 & 69.01 & 76.81 & 46.67 & 52.51 & 38.20 & \textbf{41.43} & 63.40 & 40.10 & 57.02 \\
        Mamba2 (opt / 8B)       & \textbf{67.80} & \textbf{76.17} & \textbf{70.18} & \textbf{77.78} & \textbf{47.95} & \textbf{54.61} & \textbf{40.20} & 36.88 & \textbf{69.80} & \textbf{40.86} & \textbf{58.22} \\
        \bottomrule
    \end{tabular}
\end{table}

\begin{table}[htbp]
    \centering
    \caption{Evaluation on LongBench and Passkey tasks for standard distillation.
    \textbf{Needle} denotes the exact match rate up to 32K words which is converted to roughly 42K tokens with the Llama tokenizer.
    }
    \label{tab:longbench_distill}
    \setlength{\tabcolsep}{3pt}
    \begin{tabular}{@{}l *{8}{c}@{}}
      \toprule
      \textbf{Model} & Single. QA & Multi. QA & Summ. & Few-shot & Psg ret. & Code & \textbf{LB Avg.} & \textbf{Needle} \\
      \midrule
      Mamba (eql / 70B)   & 30.83 & \textbf{32.08} & 25.42 & \textbf{62.29} & 4.5 & \textbf{49.08} & \textbf{36.97} & \textbf{70.7}  \\
      Mamba (opt / 8B)    & \textbf{31.60} & 31.28 & \textbf{25.93} & 60.73 & 4.5 & 48.52 & 36.68 & 42  \\
      \midrule
      Mamba2 (eql / 70B)  & 30.32 & 26.71 & 25.30 & 59.21 & 4.5 & \textbf{47.96} & 35.0 & 56.3  \\
      Mamba2 (opt / 8B)   & \textbf{33.14} & \textbf{32.56} & \textbf{26.37} & \textbf{64.40} & \textbf{24.5} & 45.36 & \textbf{38.98} &  \textbf{74.3}    \\
      \bottomrule
    \end{tabular}
\end{table}

The superior performance of RAD models is particularly evident in reasoning tasks, as detailed in Table \ref{tab:reasoning_distill}. Most notably, RAD Mamba2 (opt / 8B), distilled from the 8B teacher, achieves a GSM8K score of 71.27 and a CRUX score of 28.25. These results significantly surpass those of the Mamba2 (eql / 70B) baseline (GSM8K: 46.17, CRUX: 22.75), which was distilled from the much larger 70B teacher. 
This highlights RAD's ability to achieve remarkable performance gains even with a substantially smaller teacher model for distillation.

\begin{wraptable}[10]{r}{0.55\textwidth}
    \centering
    \vspace{-2\baselineskip}
    \caption{Performance on reasoning tasks (0-shot). 
    Scores are accuracy. 
    Values in parentheses denote average reasoning length ratio for each baseline (eql / 70B).
    }
    \label{tab:reasoning_distill}
    \setlength{\tabcolsep}{4pt} 
    \begin{tabular}{@{}lcc@{}}
        \toprule
        Model                 & GSM8K                 & CRUX  \\
        \midrule
        Mamba (eql / 70B)     & 62.32 (1.0)             & 26.12 (1.0)  \\
        Mamba (opt / 8B)      & \textbf{63.46} (1.36) & \textbf{26.25} (2.58)  \\
        \midrule
        Mamba2 (eql / 70B)    & 46.17 (1.0)          & 22.75 (1.0) \\
        Mamba2 (opt / 8B)     & \textbf{71.27} (2.04) & \textbf{28.25} (2.39)  \\
        \bottomrule
    \end{tabular}
\end{wraptable}

\paragraph{Ablation Study on Layer Selection.}
To demonstrate that the performance gains of RAD are indeed attributable to our principled layer selection strategy, we conduct an ablation study. 
We compare our "opt" layer selection against a "worse" layer selection. 
The "worse" set comprises layers identified by performing a reverse optimization of \cref{eq:discrete_opt_rad_en} (i.e., minimizing throughput $\tau$), using the "opt" solution as an initial exploration seed. 
This "worse" set thus represents layers whose skipping is identified as likely to be among the most \textbf{detrimental} to the throughput of self-speculative decoding, implying these layers are important to retain for efficiency (i.e., they are \textit{not} computationally redundant).

\cref{fig:self_distill_loss} illustrates the training loss curves during self-distillation for models where 8 or 14 attention layers were replaced by Mamba, Mamba2, and Longhorn, selected using either the "opt" or "worse" criteria. 
A clear divergence in training loss is observable, with the "opt" strategy consistently leading to lower loss compared to the "worse" strategy across all SSM variants.
This difference in training dynamics translates to downstream task performance. 
\cref{tab:layer_selection} summarizes the performance for the Longhorn case on key benchmarks when using "opt" versus "worse" layer selections. 
This significant performance degradation when replacing "worse" layers underscores the importance of our redundancy identification method.
 
\begin{table}[t]
    \centering
    \caption{Performance of RAD models with different attention layer selections. 
    The table shows the performance of the RAD model with Longhorn blocks, 
    where 8 attention layers are replaced with either the optimized (opt) or worse-performing (worse) selection.
    }
    \label{tab:layer_selection}
    \setlength{\tabcolsep}{3pt}
    \begin{tabular}{@{}lcccc@{}}
        \toprule
        Model            & MMLU & GSM8K & Needle (64K words) & LB Avg. \\
        \midrule
        Longhorn (opt)   & \textbf{58.27} & \textbf{63.31} & \textbf{99.7} & \textbf{45.03} \\
        Longhorn (worse) & 55.88 & 55.5 & 97.0 & 41.10 \\
        \bottomrule
    \end{tabular}
\end{table}

\section{Related Work}
\label{sec:related_work}

\paragraph{Structured LLM Pruning and Distillation.}
Several studies have investigated structured pruning, which selectively removes attention and FFN layers based on metrics such as layer importance (e.g., Block Importance \citep{men2024shortgpt, gromov2024unreasonable}) \citep{minitron2024, bercovich2024puzzle, chen2025streamlining}. 
Notably, \citep{blakeman2025nemotron-h} employs a similar approach during hybrid model construction, using methods akin to Neural Architecture Search (NAS) to explore candidate sets of layers for pruning. 
Furthermore, these studies often demonstrate that accuracy can be recovered through distillation, using the original (pre-pruning) model as the teacher and the pruned model as the student.
Our work shares similarities with these approaches in that they focus on layer redundancy to modify model architecture and employ distillation to maintain or recover performance. 
However, our approach is distinct from these prior works in two key aspects: 
(i) it identifies attention layers in existing Transformer models that become computationally redundant specifically during the self-speculative decoding process, and 
(ii) instead of merely pruning these identified layers, it replaces them with lower-cost State Space Model (SSM) layers to efficiently construct a new hybrid model.

\paragraph{Mamba is suitable for Hybrid Models?}
Several recent works \citep{lieber2024jamba, junxiongdaniele2024mambainllama} experimentally demonstrate that using Mamba in a hybrid architecture yields better results than Mamba2, especially for challenging reasoning tasks.
However, our findings indicate that Mamba2 can surpass Mamba in performance, even when distilled from a smaller teacher model, provided that the attention layers designated for replacement by SSM blocks are judiciously selected.

\section{Limitations}
\label{sec:limitations}
Our study has several limitations.
First, the Bayesian Optimization process used for solving \cref{eq:discrete_opt_rad_en} does not guarantee finding a unique or globally optimal solution, although it efficiently explores the search space.
Second, the measured throughput $\tau$, our optimization objective, can exhibit fluctuations due to the inherent non-deterministic nature of GPU computations, potentially affecting the precision of the redundancy identification.
Furthermore, our experimental conditions are not fully exhaustive due to the multitude of factors potentially affecting performance and our limited research resources.
We primarily tested replacing 50\% of attention layers; exploring different replacement ratios (either fewer or more layers) could yield further insights but was beyond the scope of this initial study.
Also, the range of models tested is limited. The use of other base LLMs (e.g., Qwen, Gemma, DeepSeek) with other SSM variants (e.g., RWKV-7 \citep{peng2025rwkv7gooseexpressivedynamic}) has not been explored, and further investigation is needed to ascertain if our results generalize to such cases.

\section{Conclusion}
\label{sec:conclusion}

This paper presented RAD (Redundancy-Aware Distillation), a novel framework that leverages self-speculative decoding to identify redundant attention layers in Transformer models.
By selectively replacing these layers with SSM blocks and performing targeted distillation, 
RAD achieves significant performance improvements in hybrid models.
We demonstrated that RAD enables student models to outperform their original teacher on reasoning tasks through self-distillation.
Furthermore, our results provide empirical evidence that RAD is effective in standard knowledge distillation setups, achieving faster convergence and higher accuracy than existing methods.

{\small
\bibliographystyle{plainnat}
\bibliography{refs}
}

%
%

\newpage
\appendix



\begin{table}[htbp]
    \centering
    \caption{Evaluation on LongBench and passkey retrieval for self-distillation setting. 
    The base model Llama-3.2-3B-Instruct is the teacher. ``w/ Mamba'', ``w/ Mamba2'', and ``w/ Longhorn'' denote RAD models where 8 identified redundant attention layers are replaced by the respective SSM blocks and then self-distilled. 
    Values in parentheses under LongBench task categories denote average input length in words. 
    The score for each task category in LongBench is the average over all tasks within that category. 
    The score of \textbf{Needle} denotes the exact match rate up to 64K words which is converted to roughly 85K tokens with the Llama tokenizer.
    Bold indicates the best score among all models in each column; underline indicates the second best.
    }
    \label{tab:longbench_self_distill}
    \setlength{\tabcolsep}{3pt}
    \vspace{0.5\baselineskip}
    \begin{tabular}{@{}l *{8}{c}@{}}
      \toprule
      \textbf{Model} & Single. QA & Multi. QA & Summ. & Few-shot & Psg ret. & Code & \textbf{LB Avg.} & \textbf{Needle} \\
        & (9.2K) & (8.4K) & (7.1K) & (6.5K) & (9.2K) & (2.7K) & - & - \\
      \midrule
      \multicolumn{9}{@{}l}{\textit{Teacher model}} \\
      Llama3.2-3B-ins & \textbf{37.31} & \textbf{35.84} & \textbf{28.17} & \textbf{67.86} & \textbf{88.50} & 53.14 & \textbf{46.82} & \textbf{100}  \\
      \midrule
      \multicolumn{3}{@{}l}{\textit{Self-Distilled (8 layers replaced)}} \\
      w/ Mamba        & 34.61 & 33.27 & 27.25 & \underline{67.19} & 68.5 & \textbf{54.74} & 44.33 & \underline{100}  \\
      w/ Mamba2       & \underline{34.69} & 31.06 & 27.05 & 63.87 & 34.5 & 50.55 & 40.37 &  82.8    \\
      w/ Longhorn     & 33.96 & \underline{35.06} & \underline{27.39} & 66.92 & \underline{76.0}   & \underline{54.70} & \underline{45.03} & 99.7 \\
      \bottomrule
    \end{tabular}
\end{table}

\appendix
\section{Additional Results}
\subsection{LongBench \& Passkey Retrieval Performance}
\label{subsec:add_res_long}

This subsection provides supplementary results for the self-distillation experiments discussed in \cref{subsec:results} of the main text. 
We present detailed performance results on the LongBench \citep{bai2024longbench} and the passkey retrieval (denoted as \textbf{Needle}).

As shown in Table \ref{tab:longbench_self_distill}, the RAD models, where 8 optimally selected attention layers were replaced by SSMs, exhibit notable performance characteristics on long-context tasks despite being self-distilled with a context length of only 2048 tokens.
Consistent with observations in the main text (\cref{subsec:results}), the RAD models (e.g., \texttt{w/ Mamba}) outperform the teacher model on the Code task (LCC, RepoBench-P).
More broadly, for most LongBench task categories, except for Psg ret. (PassageRetrieval), the performance degradation of RAD models compared to the teacher is not substantial. 
This resilience, even with limited distillation context length, suggests the robustness of our layer selection and replacement strategy. 
As detailed in the ablation studies (\cref{subsec:ablation_details}, e.g., \cref{fig:longbench_self}), selecting worse layers for replacement leads to more significant performance drops, underscoring the importance of our redundancy identification method.

The PassageRetrieval task (where the model must identify which paragraph corresponds to a given abstract, described in the textbox below) shows a more varied performance. 
Here, the Longhorn-based RAD model achieves the best performance among the RAD variants and is closest to the teacher. 
This could be attributed to Longhorn's underlying formulation as an online learning problem. 
As discussed in \citep{longhorn2024} and further explored in \cref{subsec:longhorn_detail}, Longhorn utilizes a more expressive online regression objective, $\|\mat{S}\vb{k}_t - \vb{v}_t \|^2$, for modeling key-value associations, which contrasts with the linear loss objectives in Mamba2. 
This enhanced capability for associative recall likely contributes to its improved performance on the PassageRetrieval task, even in the context of a hybrid model.

\begin{tcolorbox}[title = PassageRetrieval]
    Here are 30 paragraphs from Wikipedia, along with an abstract. Please
    determine which paragraph the abstract is from.
    \{context\}
    The following is an abstract.
    \{input\}
    Please enter the number of the paragraph that the abstract is from. The answer format must be like
    “Paragraph 1”, “Paragraph 2”, etc.
    The answer is:
\end{tcolorbox}

\subsection{Ablation Study Details}
\label{subsec:ablation_details}

\paragraph{Layer Selection \& Throughput Measurement.} 
In the main text (\cref{subsec:results}), we highlighted the importance of our layer selection strategy by comparing models built with "opt", "worse", and "eql" layer selections. 
We provide further details on these layers and quantify the throughput differences.

Our Bayesian Optimization objective function, $\mathbb{E}[\tau(\mathcal{M}_p, \{x_l\})]$, is defined as an average throughput over sample data (10 random instances from PG-19, as described in \cref{subsec:setup}). 
To quantify the empirical variability of individual throughput measurements $\tau$ for specific layer configurations $\{x_l\}$, we conducted 10 measurement trials for each selected configuration after 3 warm-up runs. 
We then calculated the mean of these 10 trials, denoted as $\langle \tau \rangle$, and their standard error. 
Time measurements were performed using the \texttt{torch.cuda.Event} API on an H100 GPU.

Table \ref{tab:layers} presents the specific layer indices identified for the "opt" and "worse" configurations in both the 8-layer and 14-layer (representing 50\% layer replacement of a 28 layer Llama3.2-3B-Instruct model) scenarios, along with their measured average throughput $\langle \tau \rangle$ and standard error. 
The "eql" configuration represents the strategy of selecting layers at equal intervals, as used in baselines.

As evident from Table \ref{tab:layers}, the "opt" layer selections consistently yield higher measured throughput $\langle \tau \rangle$ compared to the "worse" selections for both 8-layer and 14-layer configurations.  
The "eql" configuration for 14-layer replacement also shows a lower throughput than our "opt" 14-layer selection ($61.69 \pm 0.08$ vs. $69.02 \pm 0.15$), further motivating our principled approach to layer identification over simple heuristics, such as selecting layers at equal intervals.
This substantial difference in throughput, which serves as our proxy for computational redundancy, directly correlates with the downstream task performance differences reported in the main text (e.g., \cref{tab:layer_selection}), where replacing "worse" layers led to significant performance degradation.

\begin{table}[tbp]
    \centering
    \caption{Details of the layers used throughout our experiments.
    These layers are selected based on the results of the Bayesian optimization process and replaced with the SSMs.
    }
    \label{tab:layers}
    \setlength{\tabcolsep}{3pt}
    \begin{tabular}{@{}l *{2}{c}@{}}
      \toprule
       & list of layer index & $\langle \tau \rangle$ (tokens/sec) \\
      \midrule
      \multicolumn{3}{@{}l}{\textit{8 layers}} \\
      opt   & [1, 3, 8, 17, 19, 23, 25, 26] &  70.90 (0.03) \\
      worse & [0, 1, 8, 9, 15, 19, 21, 24]  &  61.81 (0.02) \\
      \midrule
      \multicolumn{3}{@{}l}{\textit{14 layers}} \\
      opt    & [1, 2, 3, 5, 7, 8, 10, 17, 19, 20, 21, 22, 23, 24]  & 69.02 (0.15) \\
      worse  & [3, 4, 5, 7, 8, 9, 10, 12, 17, 21, 22, 23, 25, 26]  & 66.31 (0.10) \\
      eql    & [0, 2, 4, 6, 8, 10, 12, 14, 16, 18, 20, 22, 24, 26] & 61.69 (0.08) \\
      \bottomrule
    \end{tabular}
\end{table}

\begin{figure}[tbp]
    \centering
    \includegraphics[width=0.7\linewidth]{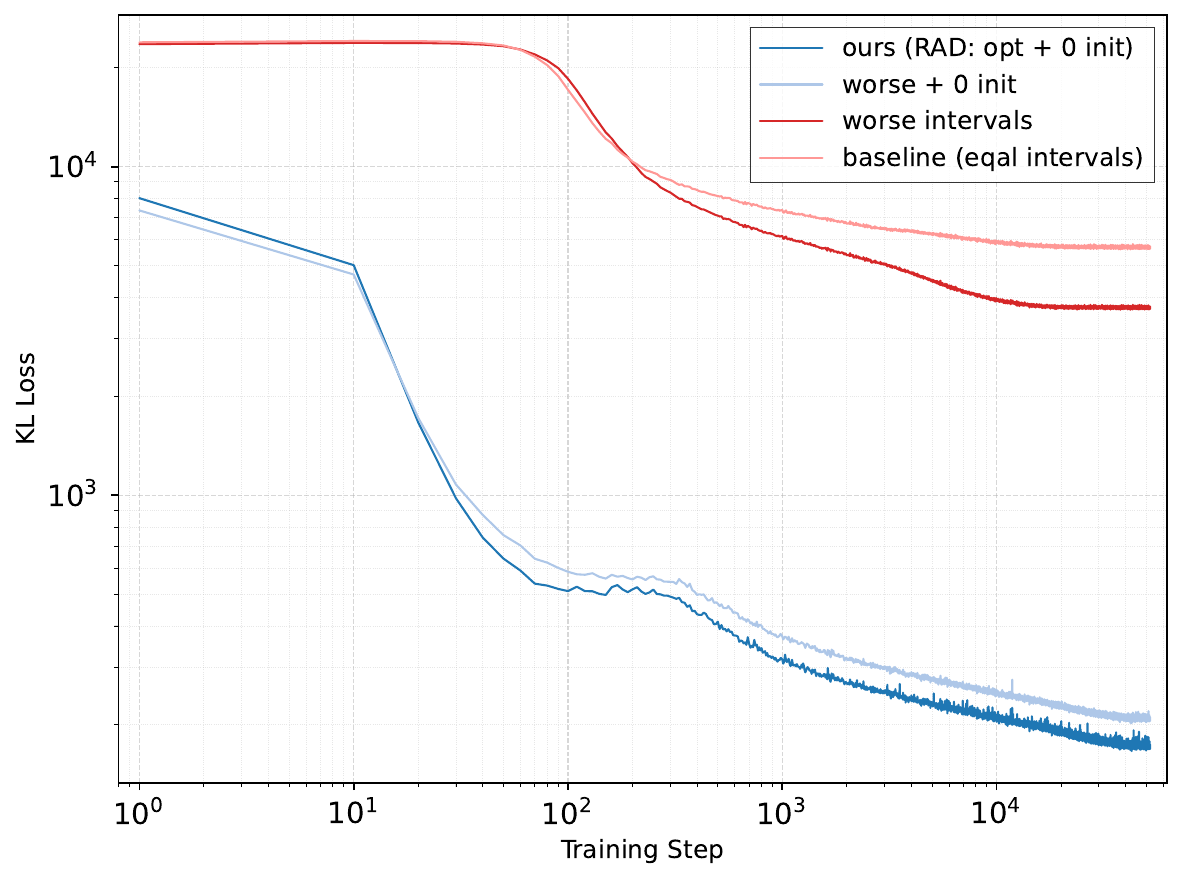}
    \caption{Comparison of self-distillation (50\% layers replaced with Mamba2) loss performance with and without zero-initialization of `in\_proj` in the SSM block.
    Optimal, worse, equal interval layer selection are also shown.
    }
    \label{fig:zero_init}
\end{figure}

\paragraph{Importance of Zero-initialization.} 
We find that the zero-initialization of the `in\_proj` layer in the replaced SSM blocks is crucial for achieving the reported good performance.
\cref{fig:zero_init} compares the self-distillation training loss curves for models where 50\% of attention layers were replaced with Mamba2 blocks, under different layer selection (detailed in \cref{tab:layers}) and `in\_proj` initialization strategies.

Specifically, we compare four configurations\footnote{
For the configurations "worse intervals" and "baseline (eqal intervals)" that do not use zero-initialization for `in\_proj`, parts of the `in\_proj` are instead initialized using weights 'qkv\_proj' from the original attention layer (mentioned in the context of baseline initialization in \cref{subsec:setup}, and illustrated conceptually in \cref{fig:rad} in main text). 
The positive effect of such initialization using original attention weights has been reported in \citep{junxiongdaniele2024mambainllama}.
}, as shown by the legend in \cref{fig:zero_init}:
\begin{itemize}
    \item \textbf{ours (RAD: opt + 0 init)}: Our proposed RAD method using optimized layer selection "opt" and the zero-initialization ("0 init") for `in\_proj`.
    \item \textbf{worse + 0 init}: Layers are selected using the "worse" criterion, but the zero-initialization for `in\_proj` is applied.
    \item \textbf{worse intervals}: Layers are selected using the "worse" criterion, and no specific zero-initialization is applied to `in\_proj` (one of the \textbf{red curves} in \cref{fig:zero_init}).
    \item \textbf{baseline (equal intervals)}: Layers are selected at equal intervals, and no specific zero-initialization is applied to `in\_proj` (another \textbf{red curve} in \cref{fig:zero_init}).
\end{itemize}

As depicted in \cref{fig:zero_init}, configurations without the zero-initialization of `in\_proj` (i.e., "worse intervals" and "baseline (equal intervals)", shown as red curves) exhibit significantly higher initial training loss and fail to decrease readily.
In contrast, applying zero-initialization (as in "worse + 0 init" and "ours (RAD: opt + 0 init)") dramatically improves the training dynamics, leading to substantially lower training loss.
Furthermore, combining zero-initialization with our optimized layer selection ("ours (RAD: opt + 0 init)") results in the most favorable loss trajectory.
This strongly suggests that initializing the replaced SSM blocks to initially mimic a "skipped" state is vital for effective distillation, and our layer selection method further optimizes this.
These results also indicate that the behavior of the training loss (in terms of its convergence value) reflects the relative magnitude of $\langle \tau \rangle$ (for the 14-layer configurations) presented in \cref{tab:layers}. 
Configurations that lead to lower throughput (such as "baseline (eqal intervals)", which has $\langle \tau \rangle = 61.69 \pm 0.08$ tokens/sec) also exhibit worse loss behavior.

\begin{figure}[tbp]
  \centering
  \begin{minipage}[t]{0.48\textwidth}
    \centering
    \includegraphics[width=0.98\linewidth]{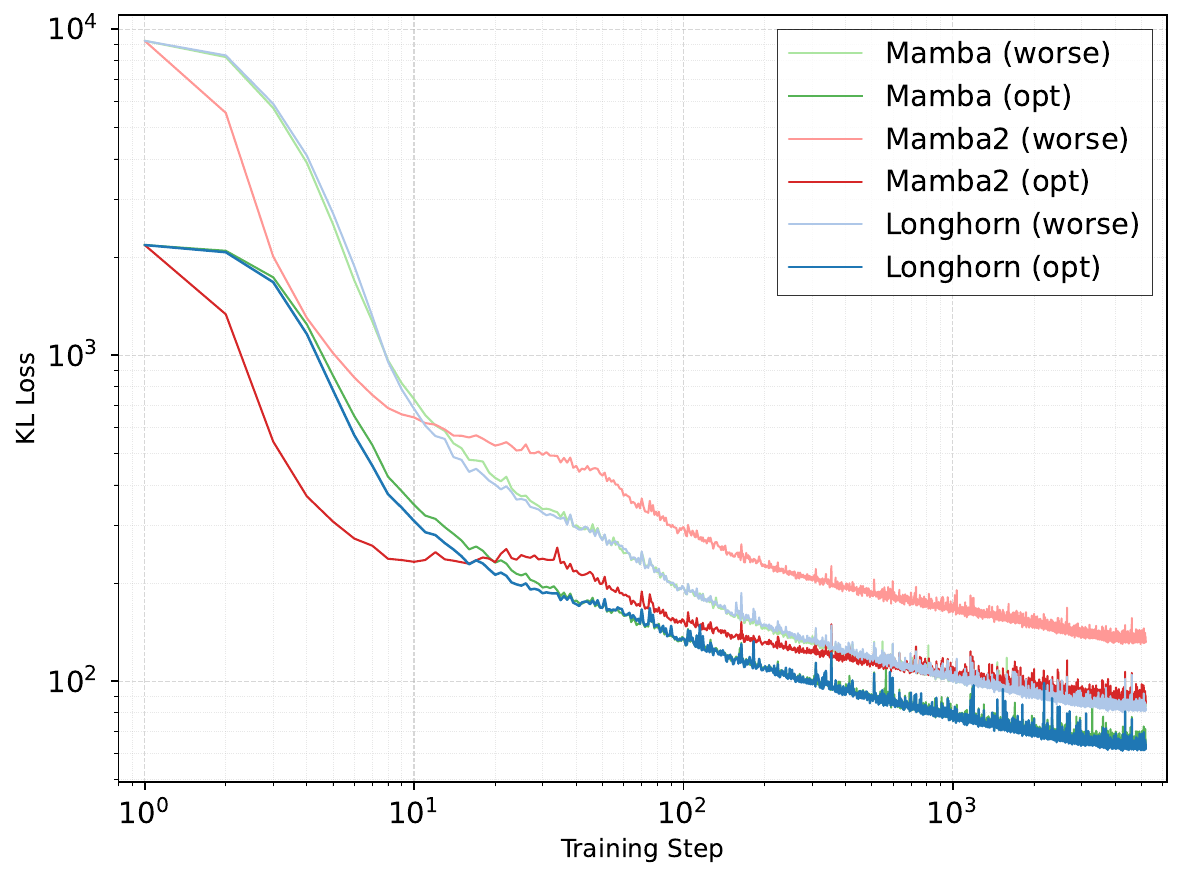}
    \caption{
    Comparison of training loss curves in the self-distillation setting, with 8 layers replaced in all models.
    }
    \label{fig:opt_vs_worse}
  \end{minipage}
  \hfill
  \begin{minipage}[t]{0.48\textwidth}
    \centering
    \setlength{\tabcolsep}{4pt}
    \vspace{-12\baselineskip}
    \captionof{table}{
    Full ablation study on reasoning tasks (0-shot) in the self-distillation setting. Eight layers are replaced with SSMs; the selected layers are listed in \cref{tab:layers}.
    }
    \label{tab:reasoning_opt_worse}
    \begin{tabular}{@{}lccc@{}}
        \toprule
        Model & layer selection  & GSM8K          & CRUX \\
        \midrule
        \multirow{2}{*}{Mamba} 
              & (worse)          & 56.79          & 24.75 \\
              & (opt)            & \textbf{62.77} & \textbf{27.38} \\
        \midrule
        \multirow{2}{*}{Mamba2}
              & (worse)          & 43.29          & 17.88 \\
              & (opt)            & \textbf{64.22} & \textbf{26.62} \\
        \midrule
        \multirow{2}{*}{Longhorn}
              & (worse)          & 55.50           & 26.50 \\
              & (opt)            & \textbf{63.31} & \textbf{27.62} \\
        \bottomrule
    \end{tabular}
  \end{minipage}
\end{figure}

\paragraph{Training Loss Curves (Self-Distillation).}
In the subsequent paragraphs, we further detail the performance differences between "opt" and "worse" layer selections observed in the self-distillation setting, as mentioned in the main text's ablation study (\cref{subsec:results}). 
We focus on training dynamics and downstream task performance to substantiate the impact of our layer selection strategy.
\cref{fig:opt_vs_worse} illustrates the training loss curves for self-distillation with 8 layers replaced by Mamba, Mamba2, and Longhorn blocks, using either the "opt" or "worse" layer selection criteria. 
Across all three SSM variants, the "opt" strategy consistently yields lower training loss than the "worse" strategy, indicating that our layer selection via self-speculative decoding effectively identifies layers more amenable to distillation.

\paragraph{Reasoning Performance.}
This difference in training dynamics translates to downstream task performance, as shown in \cref{tab:reasoning_opt_worse} for mathematical (GSM8K) and code-related (CRUX) reasoning tasks. 
For all SSM types (Mamba, Mamba2, Longhorn), selecting "opt" layers leads to substantially higher scores compared to selecting "worse" layers. 
For instance, with Mamba2, the "opt" selection achieves 64.22 on GSM8K and 26.62 on CRUX, whereas the "worse" selection results in significantly lower scores of 43.29 and 17.88, respectively. 
This stark contrast underscores the importance of appropriate layer selection.

\begin{figure}[tbp]
  \centering
  \setcounter{subfigure}{0} 
  \begin{subfigure}[b]{0.45\textwidth}
    \centering
    \includegraphics[width=\textwidth]{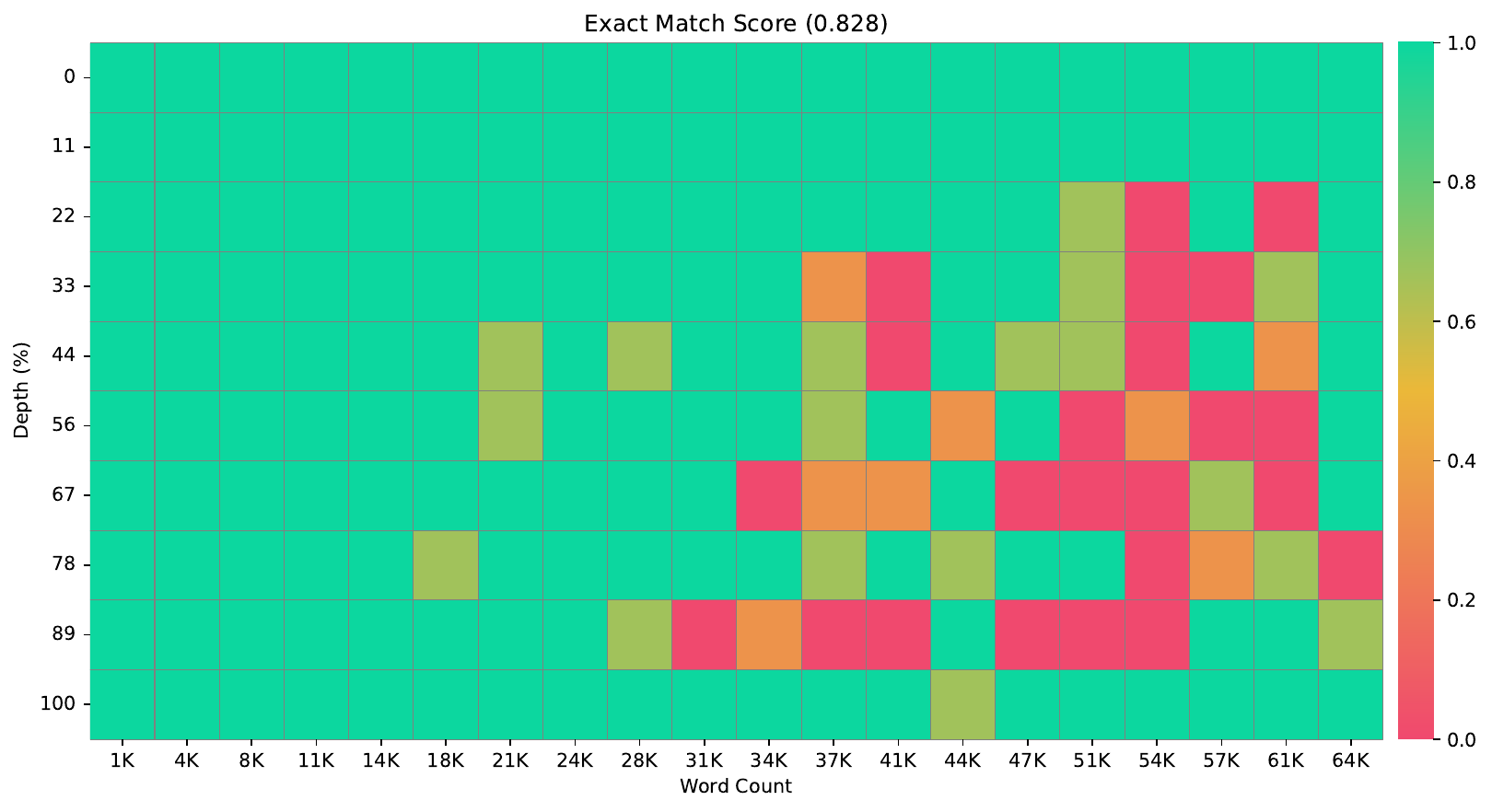}
    \subcaption{opt}
    \label{fig:pass_mamba2_opt}
  \end{subfigure}
  \hfill
  \begin{subfigure}[b]{0.45\textwidth}
    \centering
    \includegraphics[width=\textwidth]{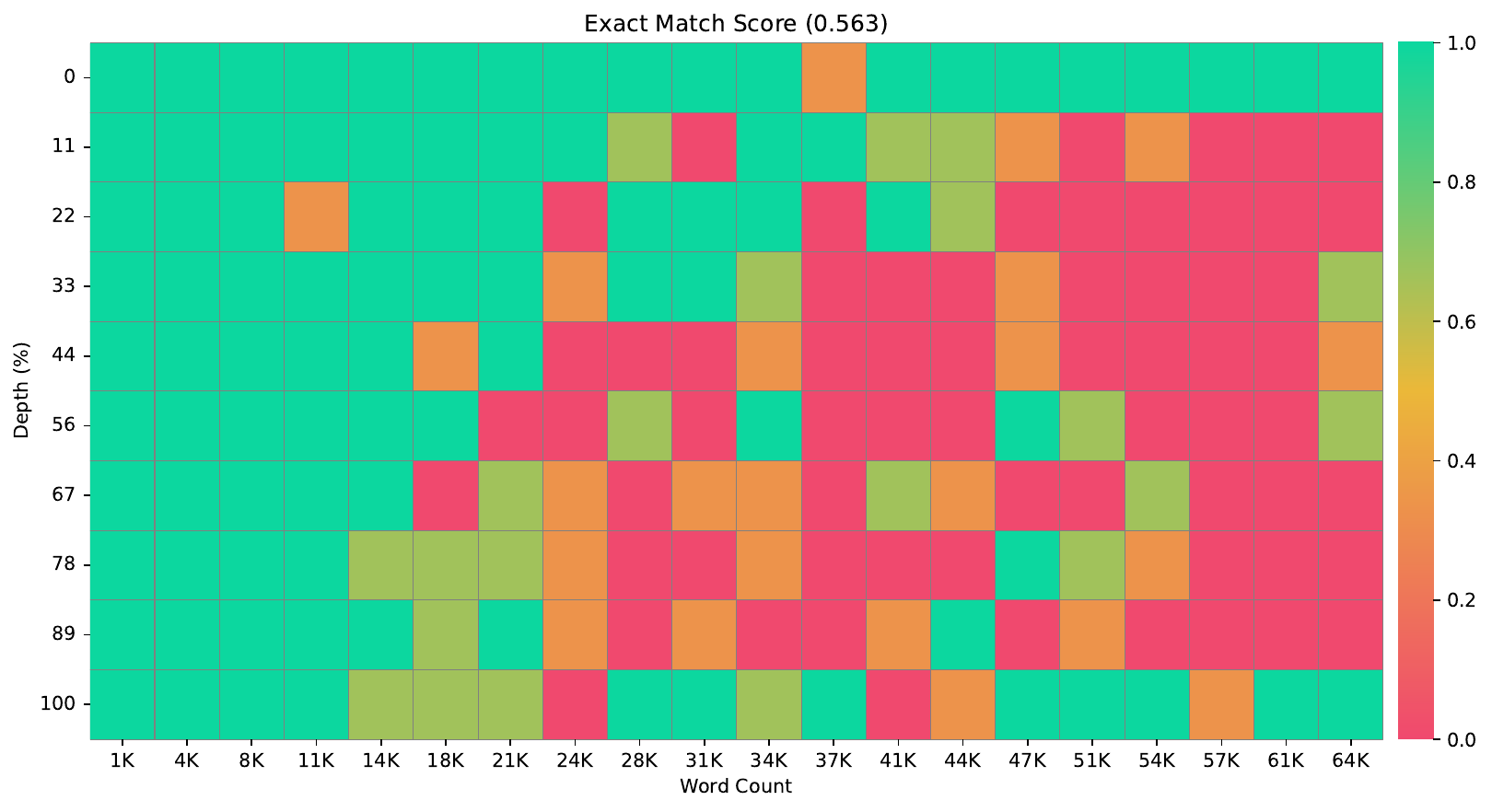}
    \subcaption{worse}
    \label{fig:pass_mamba2_worse}
  \end{subfigure}
  \paircaption{Mamba2} 
  \setcounter{subfigure}{0} 
  \begin{subfigure}[b]{0.45\textwidth}
    \centering
    \includegraphics[width=\textwidth]{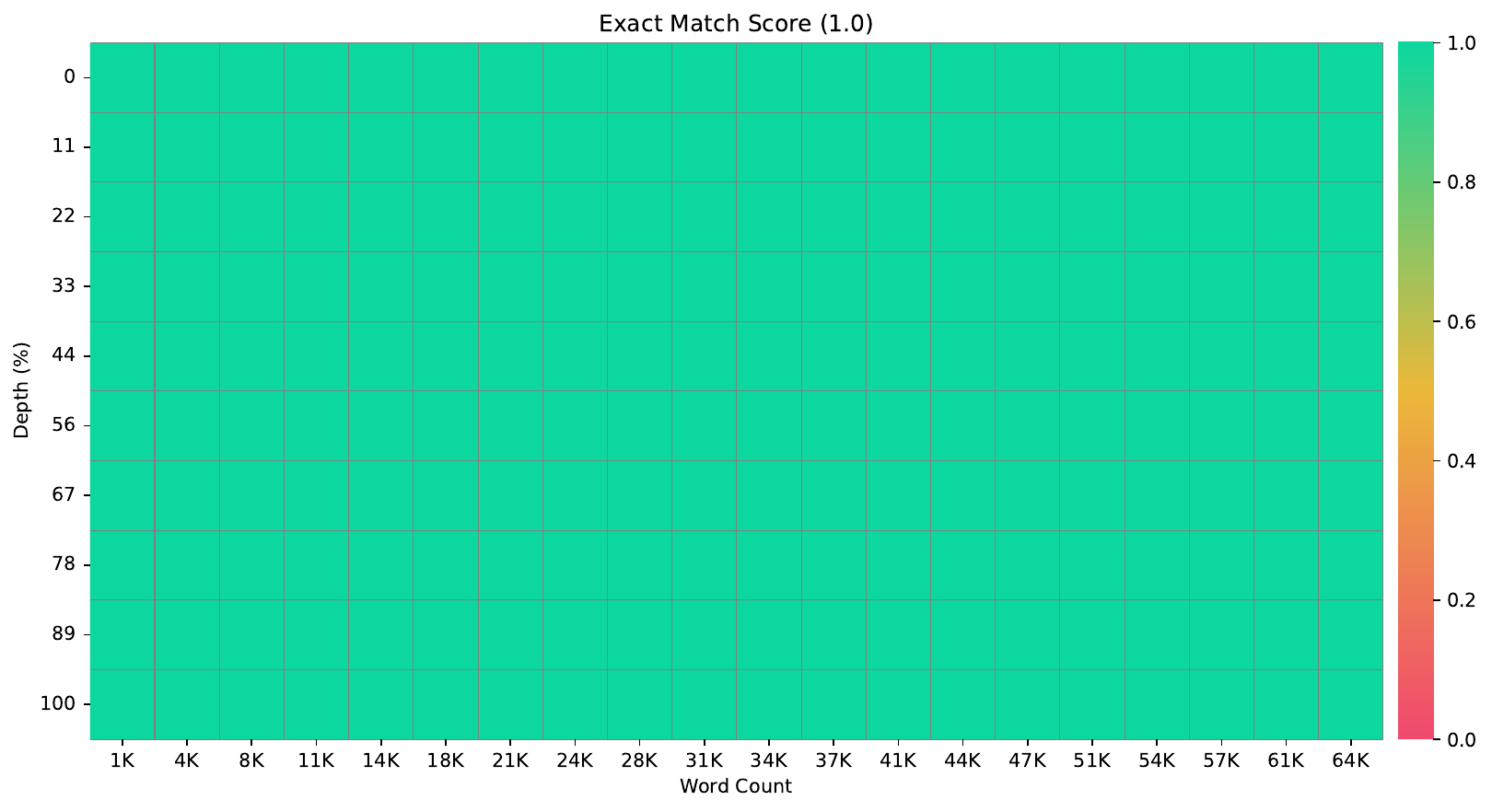}
    \subcaption{opt} 
    \label{fig:pass_mamba_opt}
  \end{subfigure}
  \hfill
  \begin{subfigure}[b]{0.45\textwidth}
    \centering
    \includegraphics[width=\textwidth]{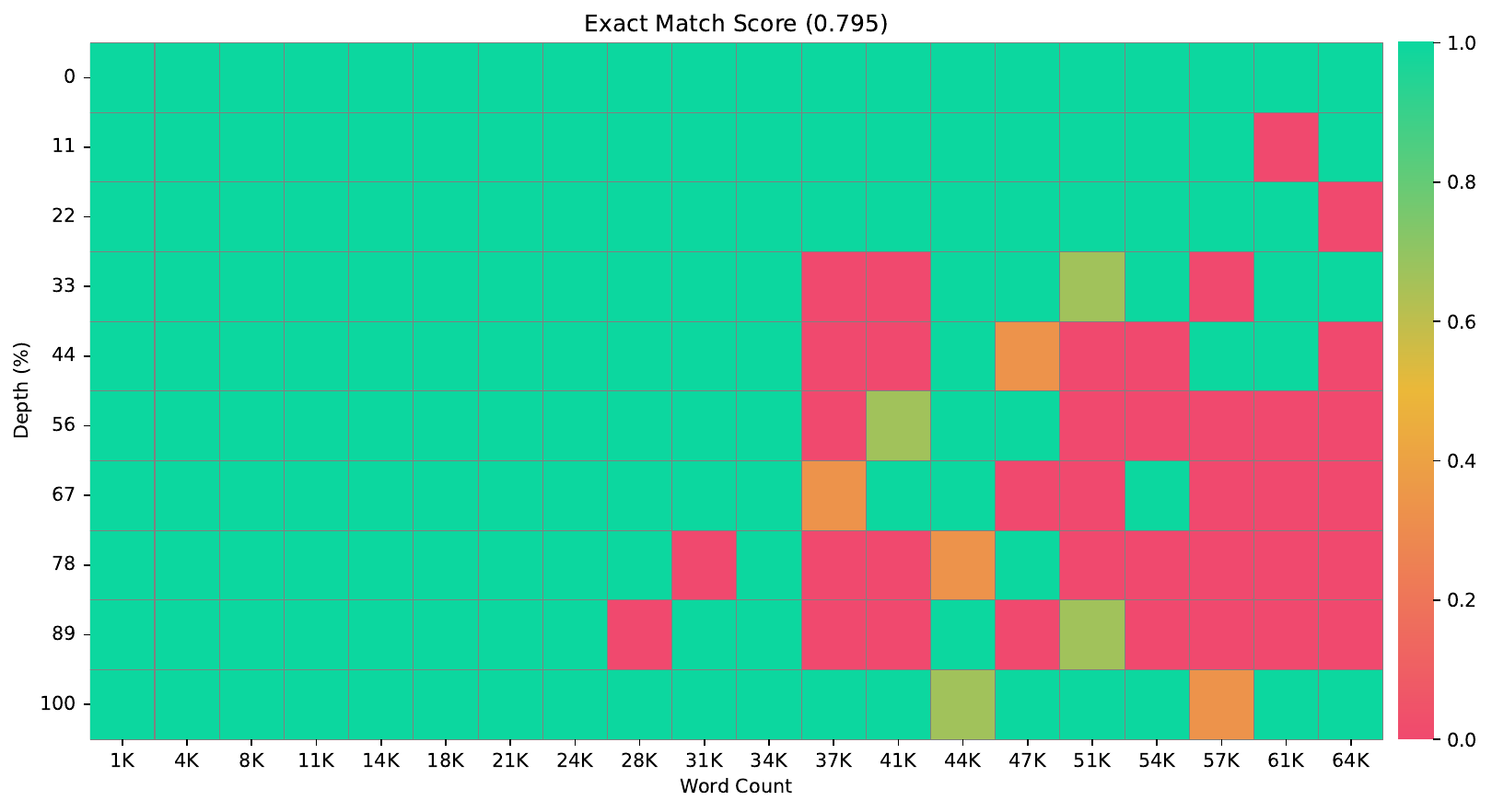}
    \subcaption{worse} 
    \label{fig:pass_mamba_worse}
  \end{subfigure}
  \paircaption{Mamba} 
  \setcounter{subfigure}{0} 
  \begin{subfigure}[b]{0.45\textwidth}
    \centering
    \includegraphics[width=\textwidth]{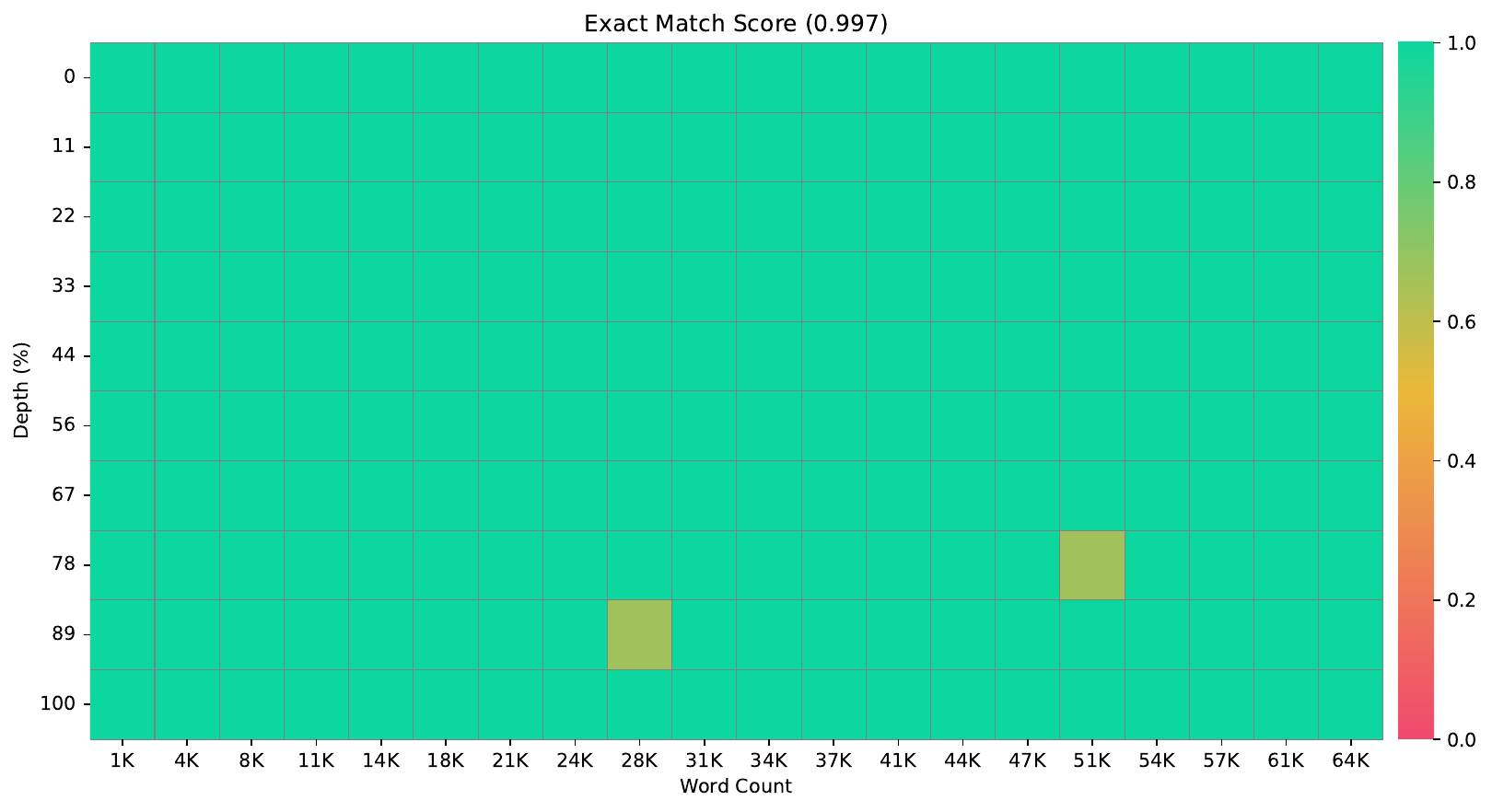}
    \subcaption{opt} 
    \label{fig:pass_longhorn_opt}
  \end{subfigure}
  \hfill
  \begin{subfigure}[b]{0.45\textwidth}
    \centering
    \includegraphics[width=\textwidth]{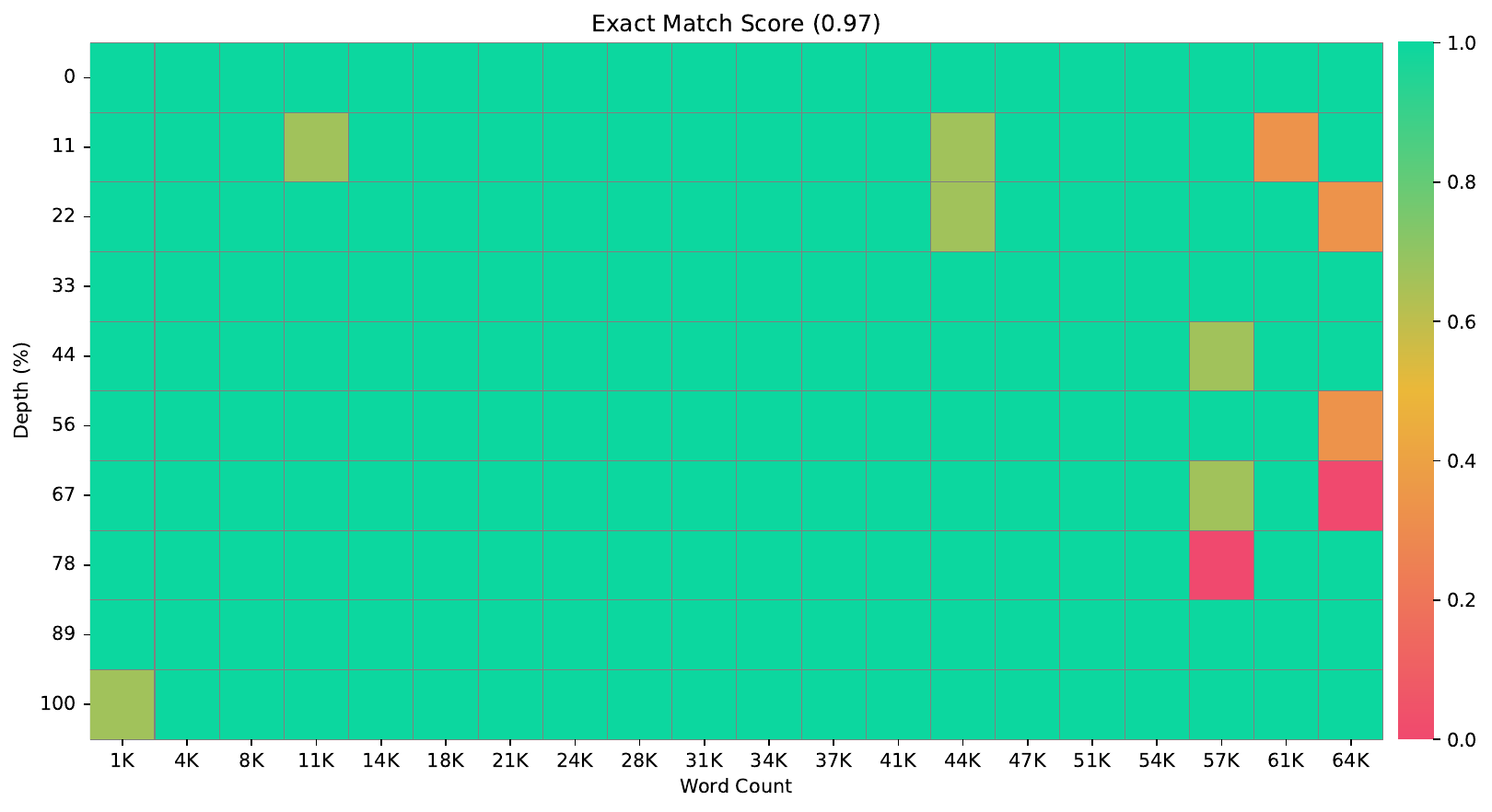}
    \subcaption{worse} 
    \label{fig:pass_longhorn_worse}
  \end{subfigure}
  \paircaption{Longhorn} 
  \caption{Passkey retrieval performance up to 64K words (roughly 85K tokens) for Llama3.2 under various model replacement configurations. Results are shown for (1) Mamba2, (2) Mamba, and (3) Longhorn, each with (a) 'opt' and (b) 'worse' layer selections. All models involved 8 replaced layers and self-distillation setting.}
  \label{fig:main_overall_passkey}
\end{figure}

\paragraph{Passkey Retrieval Performance.}
The impact of layer selection is also evident in the passkey retrieval, as shown in \cref{fig:main_overall_passkey}. 
The task setup, which involves identifying a 7-digit number embedded at varying depths within a long context (Paul Graham's essays), is described in the textbox below. 

\begin{tcolorbox}[
title = Number in a Haystack,
colframe=blue!60!white,
colback=blue!20!white,
]
    Instruction: There is an important info hidden inside a lot of irrelevant text. Find it and memorize it. I will quiz you about the important information there.

    What hard liquor, cigarettes, heroin, and crack have in common is that they’re all more concentrated forms of less addictive predecessors. Most if not all the things we describe as addictive are. [....] \textbf{The pass key is \textit{7556464}. Remember it.} And the scary [....] \\
    Query: What is the pass key? The pass key is
\end{tcolorbox}%
Each cell in the plots of \cref{fig:main_overall_passkey} represents the average exact match score computed over three samples. 
For all SSM types (Mamba2, Mamba, and Longhorn), the "opt" layer selection consistently yields better passkey retrieval performance across different context lengths compared to the "worse" selection. 
Notably, the Longhorn hybrid model appears to maintain more robust retrieval capabilities even under the "worse" layer selection strategy when compared to Mamba or Mamba2, particularly at deeper sections of the context.

\begin{figure}[tbp]
    \centering
    \includegraphics[width=\linewidth]{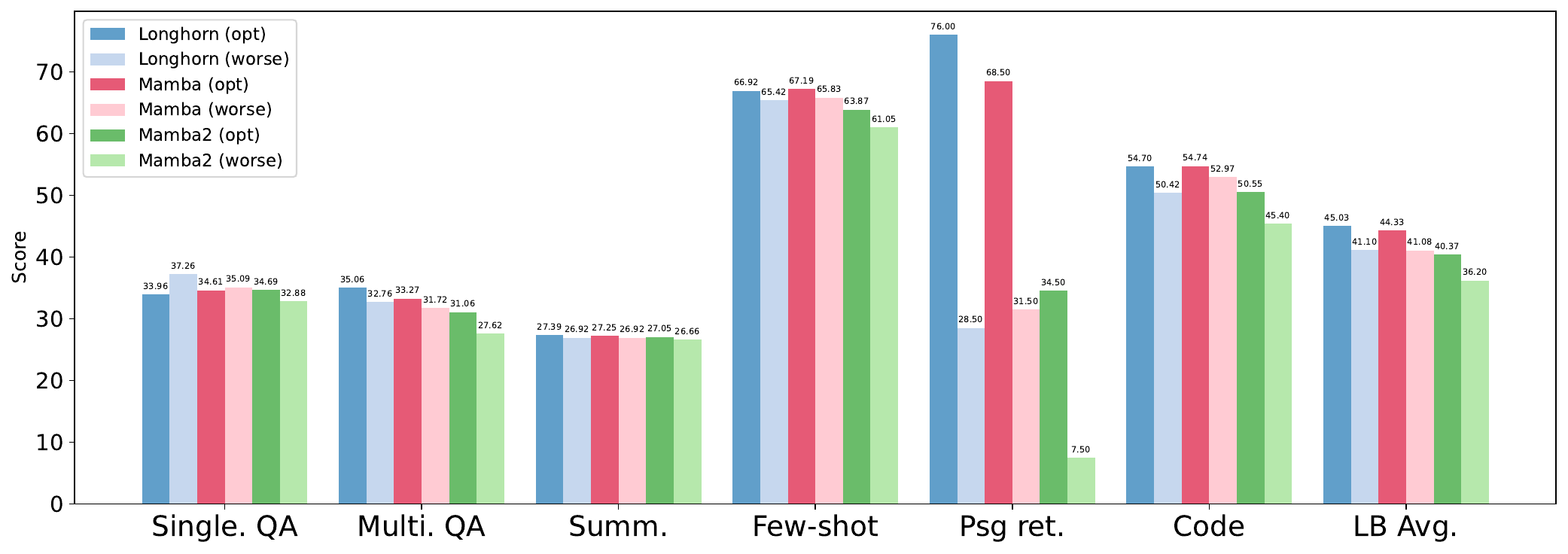}
    \caption{Full ablation study on LongBench in the self-distillation setting. 
    Eight layers (opt or worse) are replaced with Mamba, Mamba2 and Longhorn.
    Scores are averaged within each task category.
    }
    \label{fig:longbench_self}
\end{figure}

\paragraph{LongBench Performance.}
To further investigate the impact of our layer selection strategy on long-context understanding, we present a full ablation study on the LongBench in the self-distillation setting. 
\cref{fig:longbench_self} details the performance across various LongBench task categories, as described in \cref{subsec:setup} (Evaluation Metrics). 
Consistent with the trends observed in training loss dynamics and reasoning tasks, the "opt" layer selection strategy generally leads to better performance across most task categories compared to the "worse" selection for all three SSM variants.
As also noted in \cref{subsec:add_res_long}, performance on the PassageRetrieval (Psg ret.) exhibits more pronounced degradation across all RAD variants.
This indicates that performance on the Psg ret. is highly sensitive to the choice of attention layers replaced. 
Therefore, it is conceivable that if an even more optimal set of layers could be identified, further performance improvements beyond the current "opt" selection might be achievable for this particular task.

\begin{figure}[tbp]
  \centering
  \begin{subfigure}[b]{0.325\textwidth}
    \centering
    \includegraphics[width=\textwidth]{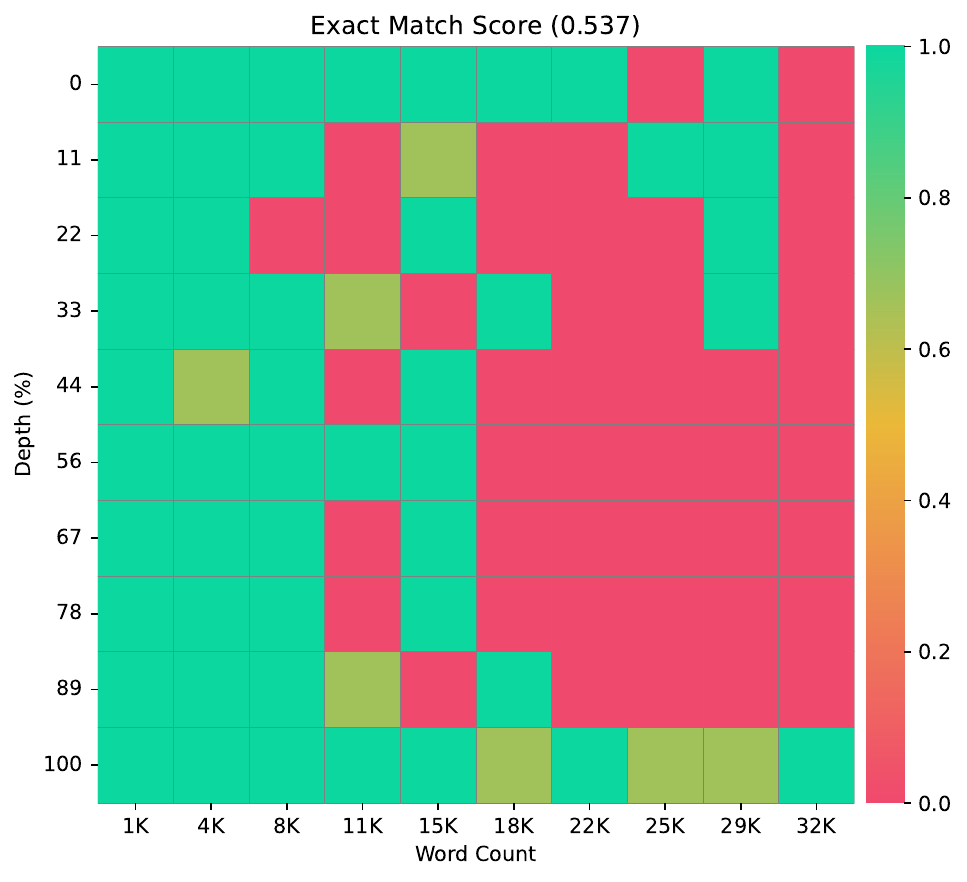}
    \caption{eql / 8B}
    \label{fig:figure1}
  \end{subfigure}
  \hfill
  \begin{subfigure}[b]{0.325\textwidth}
    \centering
    \includegraphics[width=\textwidth]{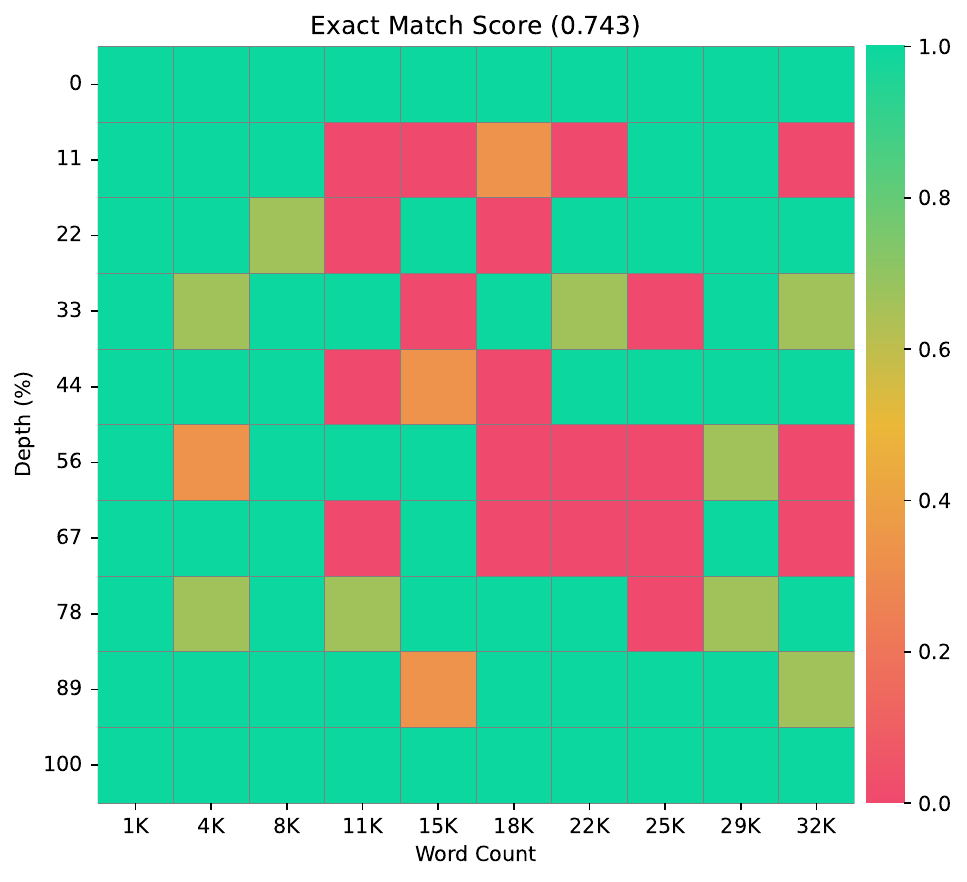}
    \caption{opt / 8B}
    \label{fig:figure2}
  \end{subfigure}
  \begin{subfigure}[b]{0.325\textwidth}
    \centering
    \includegraphics[width=\textwidth]{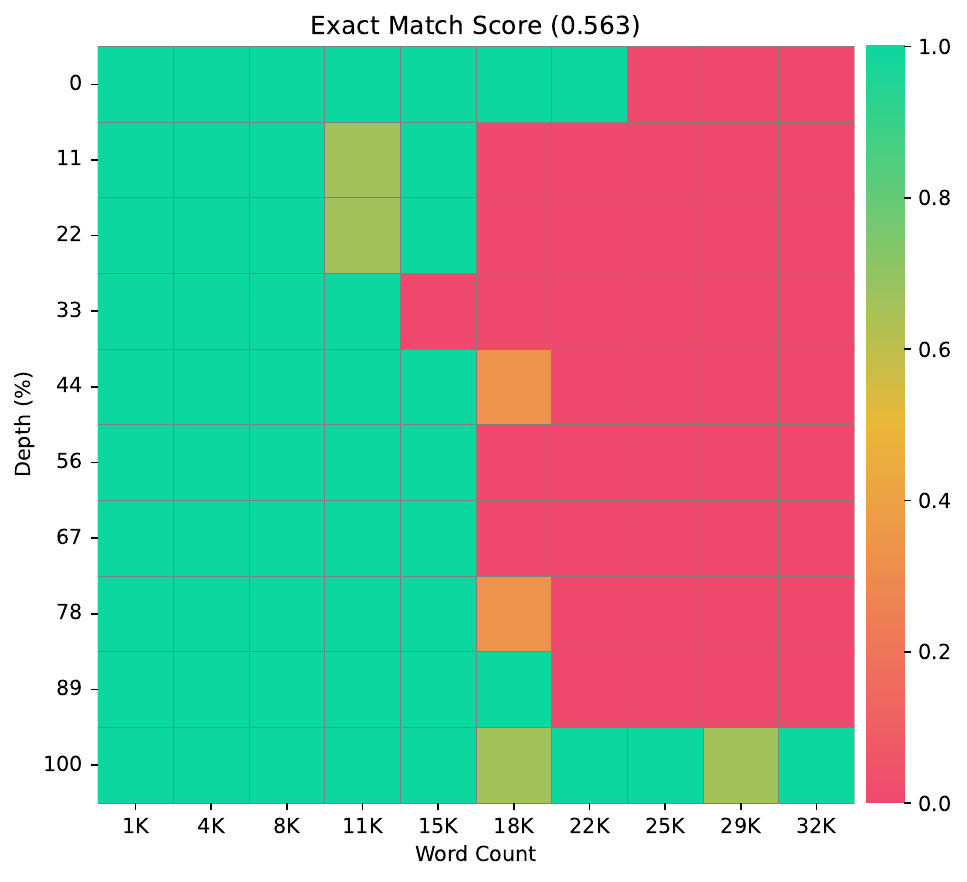}
    \caption{eql / 70B}
    \label{fig:figure3}
  \end{subfigure}  
  \caption{Passkey retrieval up to 32K words (roughly 42K tokens) for Llama3.2-Mamba2-3B (50\% layers replaced \& full-parameter distillation from large teacher models)}
  \label{fig:passkey}
\end{figure}

\paragraph{Passkey Retrieval Performance (Standard Distillation).}
We further examine the passkey retrieval capabilities in the standard distillation setting, specifically focusing on our RAD model, Mamba2 (opt / 8B). 
As established in \cref{subsec:results} (main text), this model, despite being distilled from a smaller 8B teacher, demonstrated superior performance across various tasks compared to the Mamba2 (eql / 70B) baseline, which was distilled from a much larger 70B teacher.
\cref{fig:passkey} illustrates the passkey retrieval performance of Mamba2 (opt / 8B). 
For a comprehensive comparison, we also include results for the Mamba2 (eql / 70B) baseline and an additional baseline, Mamba2 (eql / 8B), where the teacher model size is matched to our setup (i.e., distilled from an 8B teacher using the baseline's equal interval layer replacement strategy).

The results in \cref{fig:passkey} reveal several key insights.
First, comparing the two baseline models, Mamba2 (eql / 8B) and Mamba2 (eql / 70B), We observe that performance on the passkey retrieval reflects the size of the teacher model, with the 70B-distilled model achieving higher performance (53.7 for the 8B teacher vs. 56.3 for the 70B teacher).
Second, our model Mamba2 (opt / 8B), demonstrates a clear improvement in passkey retrieval across a wide range of context lengths compared to the Mamba2 (eql / 8B) baseline, which uses the same 8B teacher.

\section{Longhorn}
\subsection{Online Learning Representation}
\label{subsec:longhorn_detail}
This section provides a brief introduction to the Longhorn architecture \citep{longhorn2024}, which plays an important role in this paper. As mentioned in the main text, the recurrent form of State Space Models (SSMs) can be interpreted as implicit solvers for online learning problems.

Let $\mat{S}_t$ denote the state matrix at time step $t$. The update rule for this state matrix can be expressed as follows:
\begin{equation}
\mat{S}_t = \argmin_{\mat{S}} \mathcal{L}(\mat{S}; \mat{S}_{t-1}),
\label{eq:longhorn-state-update}
\end{equation}
where $\mathcal{L}(\cdot, \cdot)$ is an online loss function consisting of a smoothness term (e.g., $\|\mat{S} - \mat{S}_{t-1}\|$) that prevents abrupt changes in the state matrix, and a regularization term applied to $\mat{S}$.
Specifically, Longhorn formulates this online learning problem by minimizing the following loss function:
\begin{equation}
\mathcal{L}_\text{Longhorn}(\mat{S}; \mat{S}_{t-1}) =
\|\mat{S} - \mat{S}_{t-1}\|^2_\frobenius
+
\|\mat{S}\vb{k}_t - \vb{v}_t \|_{\diag(\vb{\beta}_t)}^2,
\label{eq:longhorn-obj}
\end{equation}
where $\| \mat{X} \|_\frobenius^2 = \Tr (\mat{X}^\transpose\mat{X}) = \sum_{ij} x_{ij}^2$ denotes the Frobenius norm, $\|\vb{x}\|_{\diag(\vb{\beta})}^2 = \vb{x}^\transpose\diag(\vb{\beta})\vb{x} = \sum_i \beta_i x_i^2$ is a weighted L2 norm, ${(\vb{k}_t, \vb{v}_t)}$ represents the incoming stream of key-value pairs, and $0 < \vb{\beta}_t < 1$ functions as a gating mechanism at each time step $t$.

If the first term on the r.h.s. is omitted, implying that the state matrix $\mat{S}_t$ disregards historical states prior to step $t$, the optimal solution satisfies $\vb{v}_t \approx \mat{S}_t \vb{k}_t$. This means the model directly retrieves the current value vector $\vb{v}_t$ using the key vector $\vb{k}_t$, without relying on past information.
Conversely, if the second term is omitted, the state matrix $\mat{S}_t$ remains unchanged from the previous step, completely ignoring the current observation.
By considering both terms together, the solution for $\mat{S}_t$ balances recalling current input with maintaining proximity to the previous state $\mat{S}_{t-1}$. Consequently, evaluating $\mat{S}_t \vb{k}_t$ enables the model to incorporate and retrieve diverse historical information. This mechanism can thus be interpreted as a form of associative memory.

Under moderate assumptions, \cref{eq:longhorn-state-update,eq:longhorn-obj} lead to the following explicit update formula:
\begin{equation}
\mat{S}_t =
(\mat{1} - \vb{\varepsilon}_t \otimes (\vb{k}_t \odot \vb{k}_t)) \odot \mat{S}_{t-1} + (\vb{\varepsilon}_t \odot \vb{v}_t) \otimes \vb{k}_t, \quad
\text{where} \quad \vb{\varepsilon}_t = \frac{\vb{\beta}_t}{1 + \vb{\beta}_t \|\vb{k}_t\|^2},
\end{equation}
with $\otimes$ denoting the outer product and $\odot$ the element-wise product. The detailed derivation can be found in the original paper.

Similar relationships between online learning objectives and online update formulae have been reported for other models \citep{yang2024gdn}. For example, the online learning objective for Mamba2 is given by
\begin{align}
    \mathcal{L}_\text{Mamba2}(\mat{S}) &= \|\mat{S} - \alpha_t \mat{S}_{t-1}\|_\frobenius^2 - 2  \vb{v}_t ^\transpose \mat{S}\vb{k}_t \nonumber \\
    &= \Tr[(\mat{S} - \alpha_t\mat{S}_{t-1})(\mat{S} - \alpha_t\mat{S}_{t-1})^\transpose] - 2 \Tr [\mat{S}\vb{k}_t \vb{v}_t^\transpose],
\end{align}
which yields the following explicit online update formula:
\begin{equation}
    \mat{S}_t = \alpha_t \mat{S}_{t-1} + \vb{v}_t \vb{k}^\transpose.
\end{equation}

\subsection{Longhorn's Numerical Instability} 
We observed loss spikes with Longhorn during both self- and standard distillation, indicating numerical instability. 
This instability can be addressed by applying query-key (QK) normalization, as originally proposed in \citep{henry-etal-2020-query}. 
Recently, several studies have introduced similar normalization approaches; for instance, L2 normalization for query and key vectors within SSM blocks was proposed in \citep{yang2024deltanet,yang2024gdn}.

\begin{figure}[tb]
  \centering
  \begin{subfigure}[b]{0.48\textwidth}
    \centering
    \includegraphics[scale=0.3]{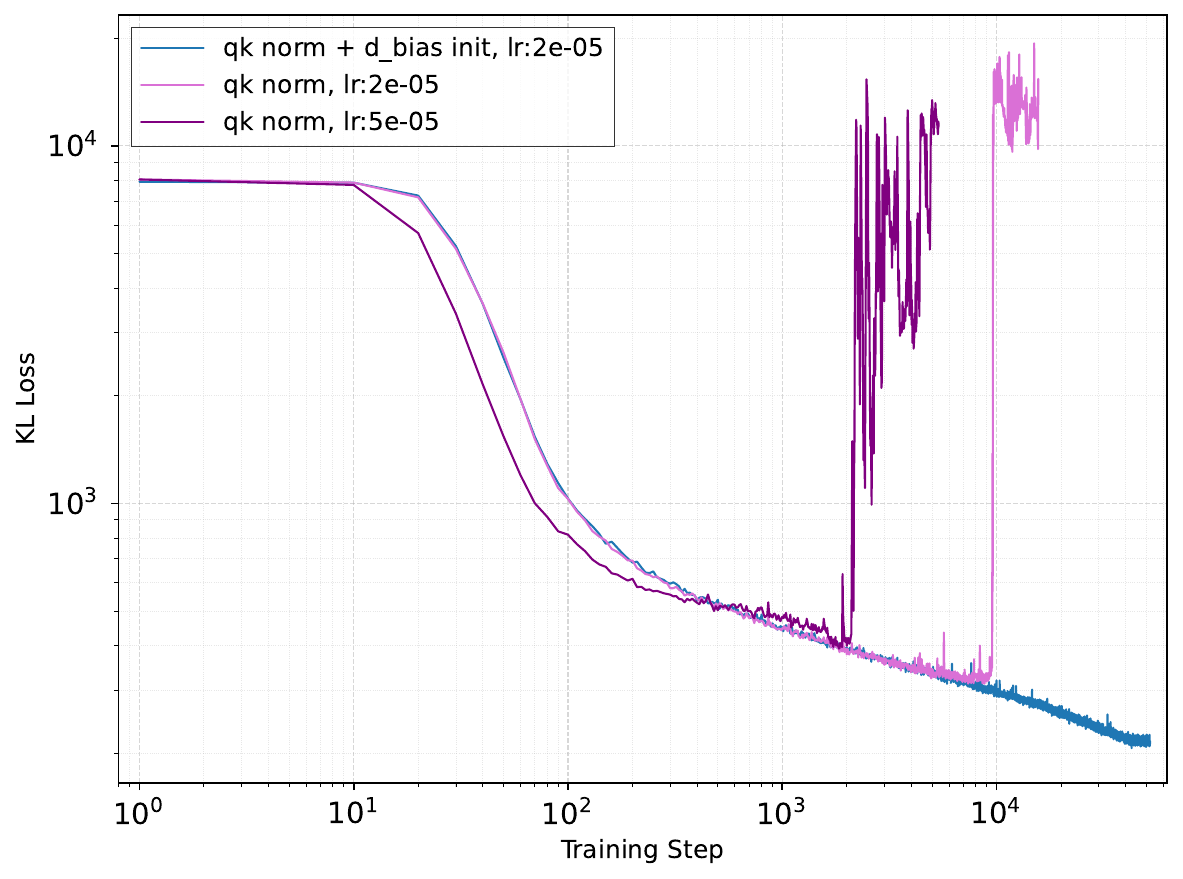} 
    \caption{loss}
    \label{fig:loss_longhorn_full}
  \end{subfigure}
  \hfill
  \begin{subfigure}[b]{0.48\textwidth}
    \centering
    \includegraphics[scale=0.3]{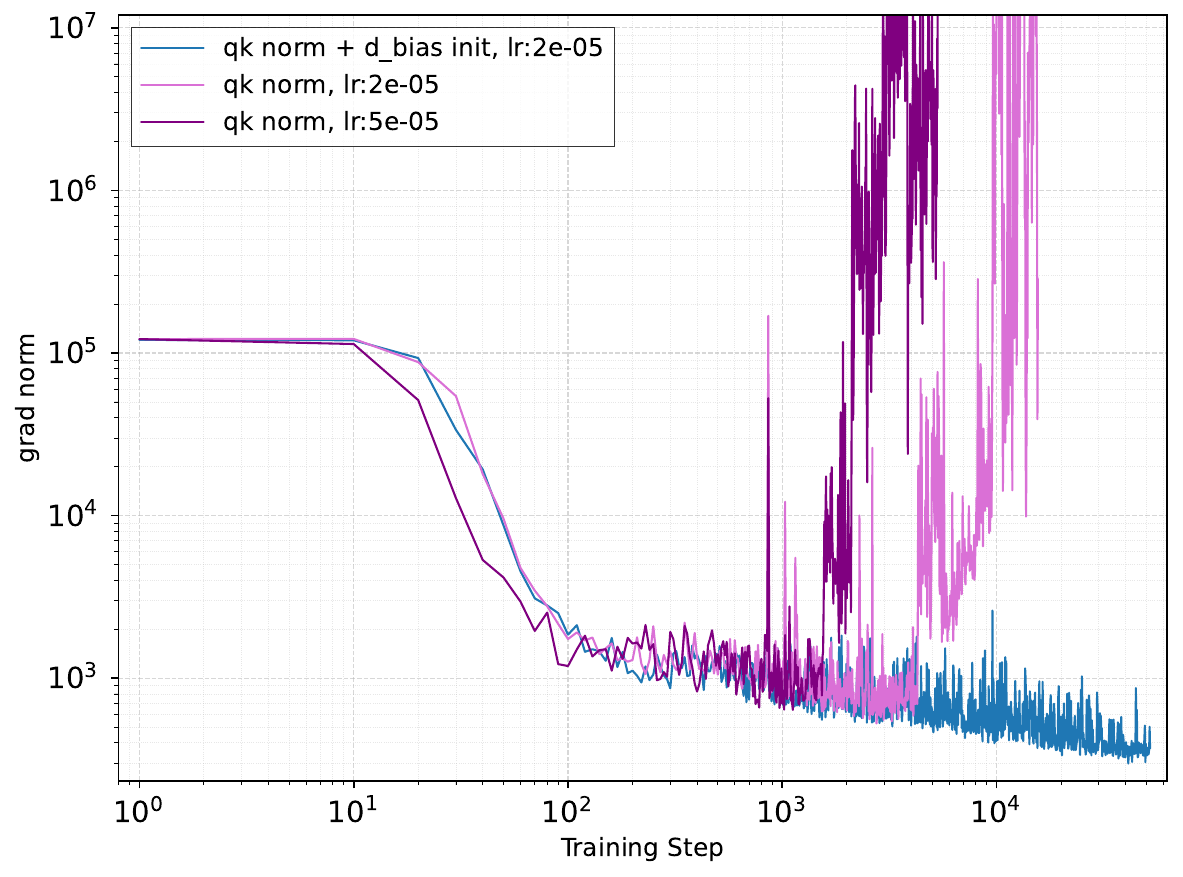}
    \caption{grad norm}
    \label{fig:loss_longhorn_gradnorm}
  \end{subfigure}
  \caption{Loss spike during standard distillation for Llama3.2-Longhorn-3B (50\% opt layers replaced)}
  \label{fig:longhorn_instability}
\end{figure}

In our experiments, we applied RMSNorm \citep{zhang2019root} to the query and key vectors in the self-distillation setting, following \citep{lieber2024jamba}. 
However, in the standard distillation setting, the instability was more acute. 
We identified that proper initialization of the $\dtbias$ parameter is crucial for stability in this context (\cref{fig:longhorn_instability}). 
Specifically, given that Longhorn's $\vb{\beta}_t$ is calculated via a sigmoid function, 
\begin{equation}
    \vb{\beta}_t = \text{Sigmoid}(\mat{W}_{\beta}\vb{x} + \text{\dtbias}),
\end{equation}
the corresponding $\dtbias$ parameter requires careful initialization. 
Suppose that $\mat{W}_{\beta}$ and $\dtbias$ are initialized with a zero-mean Gaussian distribution. In such a case, the initial values of $ \vb{\beta}_t$ tend to be centered around $\vb{\beta}_t \approx \text{Sigmoid}(0) = 0.5$. This results in relatively large initial time step sizes, which we hypothesize to be the cause of the observed numerical instability---even when the QK normalization is applied. 

Therefore, it is necessary to constrain $\vb{\beta}_t$ to relatively smaller values.
We found that, similar to Mamba's strategy for ensuring small initial time steps, $\dtbias$ should be initialized such that $\vb{\beta}_t$ falls within the range $0.001 \le \vb{\beta}_t \le 0.1$.

While training was stabilized, the standard distillation results showed no significant improvement in reasoning performance and LongBench score. A notable exception was the passkey retrieval score, which increased to 56.3 from the score of 42 for Mamba (opt / 8B) (as detailed in \cref{tab:longbench_distill} of the main text).

\section{Speculative Decoding}

In this section, we briefly describe speculative decoding \citep{leviathan2023fast, chen2023accelerating}.

Let $p(x_t \mid x_{<t})$ denote the probability distribution generated by the target model $\mathcal{M}_p$ given a prefix $x{<t}$, and let $q(x_t \mid x_{<t})$ denote the distribution generated by the draft model $M_q$. 

A token $x \sim q(\cdot \mid x_{<t})$ sampled from the draft model $M_q$ is accepted with probability 
\begin{equation}
p(\text{accept~}x) = 
\min\left(1, \frac{p(x\mid x_{<t})}{q(x\mid x_{<t})}\right).
\end{equation}
This means the token is always accepted if $q(x \mid x_{<t} ) \le p(x \mid x_{<t})$.
Conversely, if $q(x \mid x_{<t}) > p(x \mid x_{<t})$, the token is accepted with probability $p(x\mid x_{<t})/q(x\mid x_{<t})$, and 
rejected with probability 
\[
p(\text{reject~} x) = 1 - \frac{p(x\mid x_{<t})}{q(x\mid x_{<t})}.
\]
In the case of rejection, a new token is sampled from an adjusted distribution $p'(x\mid x_{<t})$ where
\begin{equation}
    p'(x\mid x_{<t}) = \frac{\max(0, p(x\mid x_{<t})-q(x\mid x_{<t}))}{\sum_x \max(0, p(x\mid x_{<t})-q(x\mid x_{<t}))}.
\end{equation}
In speculative decoding, this validation process runs for many tokens in parallel.

Crucially, this entire process ensures that the distribution of the final output sequence \emph{matches exactly} the distribution obtained by directly sampling from the target model $\mathcal{M}_p$. Hence, tokens generated through speculative decoding strictly follow the target distribution $p(x\mid x_{<t})$ (\textit{Proof}: see~\citep[Appendix]{leviathan2023fast}). Consequently, speculative decoding can be viewed as an exact sampler from the target model $\mathcal{M}_p$.

The sampling efficiency depends on the quality of draft model $q(\cdot\mid x_{<t})$.
Indeed, the overall acceptance rate $\beta$ for tokens sampled from $q(\cdot\mid x_{<t})$
is
\begin{equation}
    \beta = \sum_{x} \min(p(x\mid x_{<t}), q(x\mid x_{<t})).
\end{equation}
This acceptance rate $\beta$ quantifies how well the draft distribution $q(\cdot\mid x_{<t})$ approximates the target distribution $p(\cdot\mid x_{<t})$, and can be expressed as 
\begin{equation}
    \beta = 1 - \DTV(p(\cdot\mid x_{<t}), q(\cdot\mid x_{<t})),
\end{equation} 
where $\DTV(p, q)$ is total variation distance \citep[Thm 3.5]{leviathan2023fast}. 

\paragraph{Throughput of Speculative Decoding} In practice, the draft model typically generates $K$ tokens ($x_{t}, x_{t+1}, \dots, x_{t+K-1}$) autoregressively, and the target model validates these tokens in parallel. The acceptance rate $\beta$ actually depends on the context $x_{<t}$, but let us assume that the variance is small enough. Under these conditions, the probability that $\kappa$ tokens are accepted follows the capped geometric distribution as follows,
\begin{equation}
    p(\text{\#(accepted tokens)} = \kappa)
    = \begin{cases}
        \beta^{\kappa} ( 1 - \beta) & \text{if}\quad 0 \le \kappa < K \\
        \beta^K & \text{if}\quad \kappa = K \\
        0 & \text{otherwise}
    \end{cases}.
\end{equation}
Noting that ``$\kappa$ tokens are accepted'' implies that generation fails at the $(\kappa + 1)$-th token (which could then be corrected by the target model), the expected length of the \emph{generated} token sequence is expressed as follows,
\begin{align}
    \mathbb{E}[\text{\#(generated tokens)}] &=
    \sum_{\kappa=0}^{K}(\kappa+1)      p(\kappa) \\
    &= 
    (K+1)\beta^K + \sum_{\kappa=0}^{K-1}(\kappa+1)\beta^{\kappa}(1-\beta).
    \label{eq:generated-tokens}
\end{align}
By simply calculating \cref{eq:generated-tokens},
the expected number of generated tokens can be written as 
\begin{equation}
    \mathbb{E}[\text{\#(generated tokens)}] = \frac{1 - \beta^{K+1}}{1 - \beta} = \sum_{k=0}^{K} \beta^{k},
\end{equation} 
with each decoding step. The computation time of each step is $K\cost(q) + \cost(p)$, where $\cost(q)$ and $\cost(p)$ are the respective inference time of the draft and target models per single forward pass. Consequently, the overall expected throughput $\mathbb{E}[\tau]$ can be evaluated as 
\begin{equation}
\mathbb{E}[\tau] \approx \frac{1}{K\cost(q) + \cost(p)}\sum_{k=0}^{K} \beta^{k}.
    \label{eq:expected-throughput}
\end{equation}
Therefore, maximizing throughput primarily involves maximizing the acceptance rate $\beta$ and minimizing the computational cost of draft model $\cost(q)$, given that $\cost(p)$ cannot be easily tuned.
Also, the number of tokens $K$ that the draft model speculatively generates also matters in practice, unless $\beta \approx 1$. 

\paragraph{Interplay of TVD-Minimizing Initialization and KL Divergence in Distillation}
While our primary objective in knowledge distillation is minimizing KL divergence $\DKL(p \;\|\; q)$ between teacher $p$ and student $q$ distributions, it is a widely observed empirical principle that commencing distillation with a student model already somewhat aligned with the teacher---thereby exhibiting a smaller initial $\DKL(p \;\|\; q)$---tends to enhance learning efficiency and final performance. 

Our RAD approach aligns with this principle by employing an initialization that aims to maximize the acceptance rate $\beta$ (i.e., minimize $\DTV(p, q)$) through self-speculative decoding. 
For distributions $p$ and $q$ that are sufficiently close (i.e., $\DTV(p, q)$ is small), KL divergence is well approximated by:
\begin{equation}
    \DKL(p \;\|\; q) \approx 2 \left( \DTV(p, q) \right)^2.
    \label{eq:kl_tvd_approx}
\end{equation}
This relationship, linked to Pinsker's inequality, indicates that $\DKL(p \;\|\; q)$ scales quadratically with $\DTV(p, q)$. 
Consequently, an initialization that minimizes $\DTV(p, q)$ achieves a quadratically more substantial reduction in the initial $\DKL(p \;\|\; q)$ for the subsequent distillation stage. 

This approach of addressing the student-teacher discrepancy from the outset is also employed, for instance, in methods like TAID \citep{shing2025taid}, where dynamic interpolation between student and teacher distributions facilitates a smoother learning trajectory. 
An initialization strategy such as that employed in RAD can thus foster more efficient training, as observed in our experiments.

\section{Bayesian Optimization}
\subsection{Objective Function}
In this paper, we are interested in the case where the draft model $\mathcal{M}_p(\{x_l\})$ is build from the original model $\mathcal{M}_p$ by pruning intermediate attention layers based on the skip configuration $\{x_l\} \in \{0,1\}^L$.
In this case, the throughput \cref{eq:expected-throughput} reads
\begin{equation}
\mathbb{E}[\tau] \approx 
\frac{1}{K\cost(\mathcal{M}_p(\{x_l\})) + \cost(\mathcal{M}_p)}
\sum_{k=0}^K \left(1 - \DTV(\mathcal{M}_p(\{x_l\}), \mathcal{M}_p)\right)^k.
\label{eq:throughput-obj}
\end{equation}
We denote it as $\mathbb{E}[\tau(\mathcal{M}_p, \{x_l\})]$.
In the following optimization setup, instead of theoretically evaluate this throughput, we use actual measurements (empirical average). 
Our goal is to find the skip configuration $\{x_l\}^*$ that maximizes the expected inference throughput $\mathbb{E}[\tau(\mathcal{M}_p, \{x_l\})]$. That is, 
\begin{align}
\{x_l\}^{*} = \argmax_{\{x_l\} \in \{0,1\}^L} \mathbb{E}[\tau (\mathcal{M}_p, \{x_l\})]. \label{eq:discrete_opt_rad_en_appendix}
\end{align}

\subsection{Bayesian Optimization with Convex Relaxation}

Since exhaustively exploring all $2^L$ possible skip configurations is computationally infeasible for typical values of $L$ (e.g., $L=28$ for Llama-3.2-3B), and the relationship between $\{x_l\}$ and $\tau$ is complex as shown in \cref{eq:throughput-obj}, we employ Bayesian Optimization (BO) to efficiently search for an effective skip configuration.

Adapting the approach from \citep{zhang2023draft} for optimizing over discrete spaces, we first relax the search space $\{0,1\}^L$ to 
the convex closure $[0, 1]^L$. 
The BO algorithm then iteratively proposes points within this continuous space. To evaluate the objective function $\mathbb{E}[\tau (\mathcal{M}_p, \{x_l\})]$ for a proposed continuous point at each BO iteration, we discretize the point using a threshold of $0.5$ to obtain the corresponding binary configuration $\{x_l\} \in \{0,1\}^L$. 

The measured throughput for this discrete configuration is then used to update the BO's internal surrogate model (e.g., a Gaussian Process), guiding the subsequent search within the relaxed continuous space.

\subsection{Bayesian Optimization with constraints}

\begin{algorithm}[t]
\caption{Bayesian Optimization for Constrained $k$-Skip Configuration}
\label{alg:bo_constrained_skip}
\begin{algorithmic}[1] 
\Requirement{
Total number of candidate layers $L$\\
Number of layers to skip $k$; $(0 \le k \le L)$\\
Number of BO iterations $N_{BO}$\\
Base model $\mathcal{M}_p$
}
\Initialization{
Initialize surrogate model $\mathcal{S}$ (e.g., Gaussian Process) over the domain $[0,1]^L$.
Initialize set of all evaluated observations $\mathcal{D}_{obs} = \emptyset$.
}
\For{$i = 1$ to $N_{BO}$}
    \State 
    \algmultiline{Propose a new point $\mathbf{z}^{(i)} = (z_1^{(i)}, \ldots, z_L^{(i)}) \in [0,1]^L$ by optimizing an acquisition function derived from $\mathcal{S}$.}

    \State \textbf{Convert continuous $\mathbf{z}^{(i)}$ to discrete $\{x_l^{(i)}\}$ using top-k selection:}
    \Scope
        \State $\textit{$k$-th-largest} \gets \text{quickselect}(\{z_l^{(i)}\}, k)
        $
        \Comment{Find the $k$-th largest value}

        \For{$l = 1$ to $L$}
            \If{$z_l^{(i)} \ge \textit{$k$-th-largest}$}
                \State $x_l^{(i)} \gets 1$
            \Else
                \State $x_l^{(i)} \gets 0$
            \EndIf
        \EndFor \Comment{Now, $\{x_l^{(i)}\}$ satisfies $\sum_l x_l^{(i)} = k$}

    \EndScope

    \State  $\tau^{(i)} \leftarrow \mathbb{E}[\tau(\mathcal{M}_p, \{x_l^{(i)}\})]$.
    \Comment{Evaluate the objective function}

    \State $\mathcal{D}_{obs} \gets \mathcal{D}_{obs} \cup \{(\mathbf{z}^{(i)}, \{x_l^{(i)}\}, \tau^{(i)})\}$.
    \Comment{Add the observation}
    
    \State Update the surrogate model $\mathcal{S}$ using the pair $(\mathbf{z}^{(i)}, \tau^{(i)})$.
\EndFor
\State Select the configuration $\{x_l\}^*$ from $\mathcal{D}_{obs}$ that corresponds the highest throughput $\tau^*$.
\State \Return $\{x_l\}^*$
\end{algorithmic}
\end{algorithm}

In \cref{subsec:results} of the main text, for standard distillation, we utilized a baseline model where 50\% of the total attention layers were replaced with SSMs (e.g., \texttt{Llama3.2-Mamba-3B-distill}).
Furthermore, our ablation studies (\cref{subsec:results} and detailed in \cref{subsec:ablation_details}) investigated the impact of selecting different combinations of layers when a fixed number of layers (e.g., 8 layers) were replaced in the context of self-distillation.

To ensure a fair and rigorous evaluation of our RAD's effectiveness in identifying the optimal layers for replacement, it is crucial to assess performance differences based on which specific layers are chosen, while keeping the total number of skipped layers constant (denoted as $k$).
This allows us to separate the impact of layer selection from that of the overall attention layer pruning rate.
Consequently, we impose constraints on \cref{eq:discrete_opt_rad_en_appendix} as follows:

\begin{equation}
\begin{split}
&\{x_l\}^{*} = \operatorname*{argmax}_{\{x_l\} \in \{0,1\}^L} \mathbb{E}[\tau (\mathcal{M}_p, \{x_l\})] \label{eq:bo_const} \\
&\text{subject to} \quad \sum_{l=1}^L x_l = k 
\end{split}
\end{equation}
To solve the constrained discrete optimization problem defined in \cref{eq:bo_const}, where exactly $k$ attention layers must be selected for skipping (i.e., $\sum_l x_l = k$), several strategies can be considered. 
The discrete nature of the search space, combined with this cardinality constraint, makes the problem particularly challenging.

In our work, we adopt a strategy that integrates seamlessly with Bayesian Optimization (BO). 
Specifically, at each BO iteration, a point $\mathbf{z} = (z_1, \ldots, z_L) \in [0,1]^L$ is proposed from the continuous relaxed space by an acquisition function. To map this $\mathbf{z}$ to a valid binary configuration $\{x_l\}$ satisfying $\sum_l x_l = k$, we employ a ``top-k'' selection mechanism: the $k$ indices corresponding to the $k$ largest values in the vector $\mathbf{z}$ are identified, and their respective $x_l$ values are set to 1, while all other $x_l$ are set to 0.
This top-k approach guarantees that the constraint $\sum_l x_l = k$ is strictly satisfied at every evaluation of the objective function. 
The expected throughput $\mathbb{E}[\tau(\mathcal{M}_p, \{x_l\})]$ is then measured for this valid binary configuration $\{x_l\}$, and this observation pair $(\mathbf{z}, \mathbb{E}[\tau(\mathcal{M}_p, \{x_l\})])$ is used to update BO's internal surrogate model. 

The overall procedure, incorporating this top-k selection, is outlined in Algorithm~\ref{alg:bo_constrained_skip}. 

\section{Generation Examples}

\paragraph{Example 1}~\\
\begin{tcolorbox}[title = Question (gsm8k-main-test-\#0)]
    Question: Janet’s ducks lay 16 eggs per day. She eats three for breakfast every morning and bakes muffins for her friends every day with four. She sells the remainder at the farmers' market daily for \$2 per fresh duck egg. How much in dollars does she make every day at the farmers' market? 
\end{tcolorbox}%
\begin{minipage}[t]{.495\textwidth} 
\vspace{0pt}
\begin{tcolorbox}[title = Answer by Llama-3.2-3B-Ins]
To find out how much Janet makes at the farmers' market, we need to first calculate how many eggs she has left after eating 3 eggs for breakfast and baking 4 eggs for her friends. Then, we can multiply the remaining eggs by the price per egg to find the total amount she makes. \n{}\n{}First, calculate the total eggs used: 3 (for breakfast) + 4 (for baking) = 7 eggs. \n{}\n{}Next, subtract the total eggs used from the total eggs laid: 16 (total eggs) - 7 (eggs used) = 9 eggs. \n{}\n{}Finally, multiply the remaining eggs by the price per egg: 9 eggs * \$2 per egg = \$18.
\end{tcolorbox}%
\end{minipage}%
\hfill
\begin{minipage}[t]{.495\textwidth}
\vspace{0pt}
\begin{tcolorbox}[
    title = Answer by Self-Distilled w/ Mamba2,
    colback = green!10
]
First, calculate the total number of eggs laid per day: 16 eggs/day. Then, subtract the eggs eaten for breakfast: 16 - 3 = 13 eggs. Next, subtract the eggs used for baking: 13 - 4 = 9 eggs. Finally, multiply the remaining eggs by the selling price per egg: 9 eggs * \$2/egg = \$18/day.
\end{tcolorbox}
\end{minipage}

\paragraph{Example 2}~\\
\begin{tcolorbox}[title = gsm8k-main-test-\#17]
    Question: Jill gets paid \$20 per hour to teach and \$30 to be a cheerleading coach. If she works 50 weeks a year, 35 hours a week as a teacher and 15 hours a week as a coach, what's her annual salary?
\end{tcolorbox}%
\begin{minipage}[t]{.495\textwidth}
\vspace{0pt}
\begin{tcolorbox}[title = Answer by Llama-3.2-3B-Ins]
To find Jill's annual salary, we need to calculate her earnings from teaching and coaching separately and then add them together. First, we calculate her earnings from teaching: 35 hours/week * \$20/hour = \$700/week. Then, we calculate her earnings from coaching: 15 hours/week * \$30/hour = \$450/week. Finally, we add her weekly earnings from teaching and coaching to get her total weekly earnings: \$700/week + \$450/week = \$1150/week. To find her annual salary, we multiply her weekly earnings by the number of weeks she works in a year: \$1150/week * 50 weeks/year = \$57,500/year.
\end{tcolorbox}%
\end{minipage}%
\hfill
\begin{minipage}[t]{.495\textwidth}
\vspace{0pt}
\begin{tcolorbox}[
    title = Answer by Self-Distilled w/ Mamba2,
    colback = green!10
]
To find Jill's annual salary, we need to calculate her earnings from teaching and coaching separately and then add them together. She earns \$20 per hour as a teacher and works 35 hours a week, so her annual salary from teaching is \$20 * 35 * 50 = \$35,000. She earns \$30 per hour as a coach and works 15 hours a week, so her annual salary from coaching is \$30 * 15 * 50 = \$22,500. Adding these two amounts together gives us \$35,000 + \$22,500 = \$57,500.
\end{tcolorbox}
\end{minipage}

\paragraph{Example 3}~\\
\begin{tcolorbox}[title = crux-test-\#30]
    Question: You are given a Python function and an assertion containing an input to the function. Complete the assertion with a literal (no unsimplified expressions, no function calls) containing the output when executing the provided code on the given input, even if the function is incorrect or incomplete.\\
    
    [PYTHON]\\
    def f(array):\\
    \pyindent{1em}result = []\\
    \pyindent{1em}for elem in array:\\
    \pyindent{2em}if elem.isascii() or (isinstance(elem, int) and not str(abs(elem)).isascii()):\\            \pyindent{3em}result.append(elem)\\
    \pyindent{1em}return result\\
    assert f(["a", "b", "c"]) == ??
    
    [/PYTHON]

\end{tcolorbox}
\begin{tcolorbox}[title = Answer by Llama-3.2-3B-Ins (not correct)]
\{\\
\hspace*{1em}
\begin{tabular}{p{0.12\linewidth}p{0.8\linewidth}}
    "reasoning": &
    "The function f() filters out non-ASCII characters from the input array. It checks if an element is an ASCII character or if it is an integer and its absolute value is not an ASCII character. The function will filter out the string 'c' because it is not an ASCII character. The function will filter out the integer -123 because its absolute value is not an ASCII character. The function will filter out the string 'b' because it is not an ASCII character. The function will filter out the string 'a' because it is an ASCII character. Therefore, the output of the function will be ['a'].",\\
    "answer": &
    "[\textbackslash{}"a\textbackslash{}"]"\\
\end{tabular}\\
\}

\end{tcolorbox}
\begin{tcolorbox}[
    title = Answer by Self-Distilled w/ Mamba2 (correct),
    colback = green!10,
]
\#\# Step 1: Understand the function\\
The function `f` takes an array as input and returns a new array. It iterates over each element in the input array. If the element is an ASCII character or an integer that is not an ASCII character (i.e., its absolute value is not an ASCII character), it is added to the result array. \\

\#\# Step 2: Analyze the input\\
The input array is `["a", "b", "c"]`. We need to determine what the output of the function will be.\\

\#\# Step 3: Apply the function to the input\\
- "a" is an ASCII character, so it will be included in the result.\\
- "b" is an ASCII character, so it will be included in the result.\\
- "c" is an ASCII character, so it will be included in the result.\\

\#\# Step 4: Determine the output\\
Since all elements in the input array are ASCII characters, the output will be the same as the input array.\\

\#\# Step 5: Write the final answer\\

The final answer is the output of the function on the given input. Since all elements in the input array are ASCII characters, the output will be the same as the input array.\\

The final answer is: \\
```json\\
\{\\
\hspace*{1em}
\begin{tabular}{p{0.12\linewidth}p{0.8\linewidth}}
    "reasoning": &
     "The function will return the input array because all elements are ASCII characters.",\\
    "answer": &
    "[\textbackslash{}"a\textbackslash{}", \textbackslash{}"b\textbackslash{}", \textbackslash{}"c\textbackslash{}"]"\\     
\end{tabular}
\}
```
\end{tcolorbox}


\end{document}